\documentclass[lettersize,journal]{IEEEtran}
\usepackage{amsmath,amsfonts}
\usepackage{algorithmic}
\usepackage{algorithm}
\usepackage{array}
\usepackage[caption=false,font=normalsize,labelfont=sf,textfont=sf]{subfig}
\usepackage{textcomp}
\usepackage{stfloats}
\usepackage{url}
\usepackage{verbatim}
\usepackage{graphicx}
\usepackage{cite}
\usepackage{multicol}
\usepackage{multirow}

\usepackage{colortbl}
\usepackage{soul}
\usepackage{threeparttable}

\usepackage{threeparttable}
\usepackage{amssymb}

\usepackage{inconsolata}

\usepackage{graphicx}

\usepackage{microtype}
\usepackage{booktabs}
\usepackage[sort&compress,numbers]{natbib}
\setcitestyle{numbers,square}
\usepackage{xcolor}
\usepackage{enumitem}

\definecolor{Gray}{gray}{0.9}
\sethlcolor{Gray}

\usepackage{amsthm}          % provides theorem machinery
\theoremstyle{plain}         % default style: italic body, bold header
\newtheorem{theorem}{Theorem}

\newcommand{\bX}{\mathbf{X}}
\newcommand{\bW}{\mathbf{W}}
\newcommand{\bLma}{\mathbf{\Lambda}}
\newcommand{\bR}{\mathbf{R}}
\newcommand{\Clamp}{\mathsf{Clamp}}
\newcommand{\Wtilde}{\tilde{\mathbf{W}}}
\newcommand{\diag}{\mathsf{diag}}

\newcommand{\What}{\hat{\bW}}
\newcommand{\Map}{\mathsf{Map}}

\begin{document}

\title{Compensate Quantization Errors+: Quantized Models \\ Are Inquisitive Learners}

\author{Yifei Gao$^*$~, Jie Ou$^*$~, Lei Wang$\dag$,~\IEEEmembership{Senior Member,~IEEE,}  \\ Jun Cheng,~\IEEEmembership{Senior Member,~IEEE,} and Mengchu Zhou,~\IEEEmembership{Fellow,~IEEE}
        % <-this % stops a space
%\thanks{Y. Gao is with University College of London. E-mail: ucabaoj@ucl.ac.uk. Ou Jie is with University of Electronic Science and Technology of China. E-mail: 202211090813@std.uestc.edu.cn.}
\thanks{$*$ Equal contribution.}
\thanks{$\dag$ Corresponding author.}
\thanks{Preprint. This paper is under review in TNNLS.}
}

\markboth{IEEE Transactions on Neural Networks and Learning Systems,~Vol.~X, No.~X, April~2025}%
{Shell \MakeLowercase{\textit{et al.}}: A Sample Article Using IEEEtran.cls for IEEE Journals}


\maketitle

\begin{abstract}
The quantization of large language models (LLMs) has been a prominent research area aimed at enabling their lightweight deployment in practice. Existing research about LLM's quantization has mainly explored the interplay between weights and activations, or employing auxiliary components while neglecting the necessity of adjusting weights during quantization. Consequently, original weight distributions frequently fail to yield desired results after round-to-nearest (RTN) quantization. Even though incorporating techniques such as mixed precision and low-rank error approximation in LLM's quantization can yield improved results, they inevitably introduce additional computational overhead. On the other hand, traditional techniques for weight quantization, such as \textit{Generative Post-Training Quantization}, rely on manually tweaking weight distributions to minimize local errors, but they fall short of achieving globally optimal outcomes. Although the recently proposed \textit{Learnable Singular-value Increment} improves global weight quantization by modifying weight distributions, it disrupts the original distribution considerably. This introduces pronounced bias toward the training data and can degrade downstream task performance. In this paper, we introduce \textit{Singular-value Diagonal Expansion}, a more nuanced approach to refining weight distributions to achieve better quantization alignment. Furthermore, we introduce \textit{Cross-layer Learning} that improves overall quantization outcomes by distributing errors more evenly across layers. Our plug-and-play weight-quantization methods demonstrate substantial performance improvements over state-of-the-art approaches, including OmniQuant, DuQuant, and PrefixQuant.
\end{abstract}

\section{Introduction}

% task background
Large language models (LLMs) have attracted considerable attention due to their exceptional performance on a variety of downstream tasks~\cite{gpt4}\cite{llama}. However, training and maintaining these LLMs require substantial resources. In this scenario, quantization becomes an essential strategy, providing solutions to decrease both memory and computational requirements of LLMs.
%footprint

One of the primary challenges in quantization lies in handling outliers, which can be broadly categorized into normal and massive outliers~\cite{sun2024massive}. Traditional approaches address this issue by using such techniques as mixed precision~\cite{awq}\cite{spqr}, low-rank approximation~\cite{xu2023qalora}\cite{saha2024compressing}, and weight-activation smoothing~\cite{smoothquant}\cite{omniquant}. Mixed precision and low-rank approximation are effective in mitigating both normal and massive outliers, but they inevitably introduce additional computational overhead and cannot fully compensate for massive outliers. In contrast, smoothing can be elegantly integrated into the model itself with the minimal overhead, but it is only effective in alleviating normal outliers. Recent work has primarily focuses on addressing massive outliers~\cite{liu2024spinquant,zhao2310atom,ashkboos2024quarot}. The residual structure of LLMs, while facilitating training, leads to the dramatic amplification of activation gradients as they accumulate. However, all these efforts often overlook the impact of the model linear weights themselves. As quantization introduces errors unavoidably, simply preserving the original weight distribution through various approaches fails yield the optimal results~\cite{liu2023qllm}. As such, appropriately adjusting the original weight distribution becomes essential.

%------------------------------------------------------
\begin{figure*}[!t]
\includegraphics[width=1\textwidth]{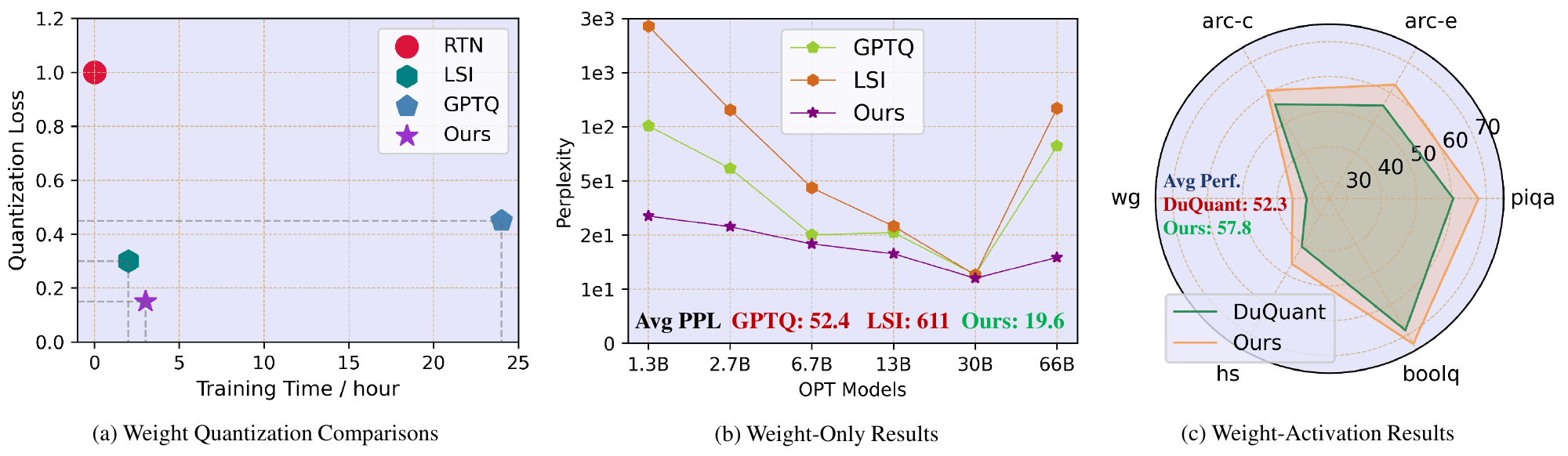}
\centering
%\vspace{-1em}
    \caption{(a) Compared to existing PTQ weight quantization methods, our techniques achieve superior loss reduction while preserving fast training times. (b) The W2A16g128 quantization results. Our methods consistently deliver superior quantization results even for extremely low-bit settings. (c) The W3A4 quantization on LLaMA-7B. Our techniques significantly outperform the original DuQuant approach, yielding substantially improved performance.}
    \label{fig:teaser}
    %\vspace{-1.5em}
\end{figure*}
%------------------------------------------------------

For linear-weight quantization in LLMs, the Generative Post-Training Quantization (GPTQ) series~\cite{frantar2022optimal}\cite{gptq} employs a handcrafted process in which individual weights with the least error after quantization are incrementally rounded to $n$-bit precision. After several steps, the Hessian matrix of the weights is updated, allowing subsequent weight adjustments. In contrast, round-to-nearest (RTN) applies a direct MinMax rounding approach, mapping weights to the nearest $n$-bit integer without additional computation. More recently, in Learnable Singular-value Incremental (LSI)~\cite{lsi}, Gao et al. introduced a novel approach by proactively redistributing the magnitudes of weight matrices to better align them with RTN calibration. This method involves decomposing weight matrices through singular value decomposition (SVD) and making the resulting diagonal singular values learnable. 

RTN employs a straightforward quantization approach, but delivers the poorest results in general. GPTQ, while offering quantization-aware weight alignments, fails to achieve global optima and demands extensive calibration time. In contrast, LSI effectively redistributes weights to better align with activations. However, it excessively adjusts the weights to accommodate activation outliers, resulting in pronounced overfitting and biased downstream task performance. These limitations in existing methods have inspired this work; our improvements over them are shown in Fig.~\ref{fig:teaser}.

In this paper, we mainly aim to address the limitations of LSI while leveraging its strengths to improve weight quantization techniques. We begin by analyzing the key factors behind LSI’s effectiveness and, building on these insights, introduce a novel approach: \textbf{S}ingular-value \textbf{D}iagonal \textbf{E}xpansion (SDE). This method refines LSI by incorporating additional learnable parameters into the diagonals adjacent to the singular-value matrix of linear weights. To further distribute quantization errors across layers evenly, we adopt \textbf{C}ross-layer \textbf{L}earning (CL) that considers errors that may propagate into subsequent layers. Both SDE and CL are designed to be plug-and-play and can be easily integrated into existing workflows.

Through extensive experiments, we have demonstrated the superiority of our weight quantization techniques across a range of tasks and various benchmarks. Specifically, for OmniQuant~\cite{omniquant}, we enable the original failed quantization results (NaN) on LLaMA-2(3)-70B quantizable on W4A4KV4 In DuQuant~\cite{lin2024duquant}, we significantly boosted PPL (4+) and downstream task performance by over 6\% for LLaMA-7B models on W3A4KV4. Furthermore, we enable the performance of Vicuna models on MT-Bench to largely surpass the original OmniQuant and DuQuant, with even a $10\%$ performance gain on the benchmarks~\cite{mmlu}\cite{bai2023longbench}. Our new contributions are summarized in as follows:
\begin{enumerate}[label=\arabic*)]
    \item In-depth analyses of the necessity of weight adjustments by redefining the quantization of weights as an inequality-solving problem.
    \item Introduction of SDE and CL to minimize quantization errors and channel the burden of quantization across layers, with theoretical proof to demonstrate the superiority of our techniques over LSI.
    \item Extensive experiments conducted across various tasks, baselines, and models, showing significant improvements in precision and exhibiting strong potential for industrial applications.
\end{enumerate}

\section{Related Work}
\subsection{KV Cache Reduction}
With the recent development of LLMs, downstream applications like multi-turn dialogue~\cite{bahrini2023chatgpt} and long-document summarization~\cite{goyal2020evaluating} have become increasingly complex and diverse. They typically require the storage of a vast number of hidden states (KV Cache) whose size can even exceed that of LLM itself. Similar to model quantization, KV Cache reduction—an approach specifically aimed at reducing the computational cost of processing long sequences—has seen significant progress in recent years.

Existing methods have exploited the inherent sparsity in the attention mechanism of LLMs. For instance, Heavy-hitter Oracle~\cite{zhang2023h2o} determines important tokens (heavy hitters) based on accumulated attention scores, discarding those unimportant tokens to enable inference with only 20-40$\%$ of the original KV Cache. MiniCache~\cite{liu2024minicache}, observing the high consistency of deep KV Caches (attributed to the residual nature of LLMs), merges adjacent layers’ KV Caches while retaining a small set of highly distinctive tokens. CaM~\cite{zhang2024cam} leverages the high cosine similarity between adjacent token KV Caches to apply token merging for memory reduction. Such methods as KVQuant~\cite{hooper2024kvquant} focus on quantizing the KV Cache to low bit-width representations.

Recent advances have built upon these foundational approaches~\cite{zhang2023h2o,dejavu,ge2023model} by combining multiple optimization strategies and offering more fine-grained observations at the head level of the attention mechanism. For example, DuoAttention~\cite{xiao2024duoattention} separates attention heads into retrieval heads and streaming heads. Dynamic Discriminative Operations~\cite{wan2024d2o} improves H2O by refining both the inter-layer KV budget allocation strategy and heavy-hitter identification process, while also adopting token merging to further reduce memory consumption. Other studies~\cite{zhang2024lorc} have applied low-rank approximation to KV Caches, supplementing more sophisticated token selection strategies to the attention mechanisms. Additionally, some methods~\cite{wuhshare} reduce computation by reusing top-k important KV Cache indices across time steps.

\subsection{Quantiztaion Methods}
\label{sec:quant_methods}
\paragraph{General Quantization Methods}
A variety of quantization methods have been developed. Quantization-Aware Training (QAT)~\citep{llmqat}\cite{chen2024efficientqat} involves adjusting model parameters during the training phase to align quantization techniques. Post-Training Quantization (PTQ)~\citep{yang2024mitigating}\cite{omniquant}, modifies models without the need for extensive training processes, making PTQ methods notably quicker than QAT. The primary difference between QAT and PTQ lies not only in training time but also in the amount of data required for training or calibration. The latter typically requires only a small amount of calibration data—for example, 128 samples of sequence length 2048 are often sufficient for calibration or lightweight fine-tuning. In contrast, QAT generally demands a significantly larger dataset, often comparable to that required for full model retraining. Moreover, due to the flexibility inherent in quantization itself, extensive global training is not always necessary. As a result, some PTQ methods~\citep{lin2024duquant}\cite{omniquant} can even surpass the results achieved by QAT ones.

\paragraph{Quantization Outlier Suppression}
Outliers in activations~\cite{chee2024quip}\cite{ma2024affinequant} primarily cause quantization errors. Historically, these outliers have often been preserved and free from quantization~\cite{spqr}\cite{awq}. This operation requires hardware accelerators~\cite{tseng2024quip} capable of mixed-precision inference. But even with specifically designed kernels, the additional consumption of memory and time is unavoidable. Scale transformation techniques~\cite{smoothquant}\cite{omniquant} are developed to address this issue by partially transferring the high-magnitude components of activations to weights. While effective for normal outliers~\cite{smoothquant}, they struggle to handle massive outliers~\cite{sun2024massive}. The smoothing of massive outliers results in magnitude collapse on weight because of their excessive salience. Recently, rotation-based transformations~\cite{liu2024spinquant}\cite{ashkboos2024quarot} have emerged as a promising solution. They can effectively mitigate both normal and massive outliers, making 4-bit quantization of weights and activations largely compatible with the original model. Additionally, since massive outliers are often specific to certain tokens, recent methods~\cite{chen2024prefixquant,son2024prefixing} have introduced token-wise pre-identification strategies. By isolating these tokens during quantization, these techniques can significantly reduce overall quantization errors.

\section{Preliminaries}
%\paragraph{Uniform Quantization}
\subsection{Uniform Quantization}
The quantization of both weights and activations in LLMs can be uniformly expressed as follows:
\begin{equation}
\label{eq:quantizer_func}
    \mathbf{X_{int}} = \mathsf{Clamp}(\lfloor \frac{\mathbf{X}}{s_{x}} \rceil + z, 0, 2^{n}-1),
\end{equation}
where $\mathbf{X_{int}}$ denotes a weight or activation matrix, $\Clamp(\cdot)$ is a rounding function, $n$ the number of designed bit, $s_{h}$ and $z$ are scaler and zero point corresponding to $\mathbf{X}$ after token-wise or channel-wise~\cite{awq} calibration, respectively. 

Transformation and rotation techniques can be integrated into weight matrix $\bW$ and activations $\bX$ for better calibration during quantization, i.e.,
\begin{equation}
\label{eq:uniform_quant}
    \bX \cdot \bW \approx Q((\bX \cdot \bLma) \bR) \cdot Q(\bR^T(\bLma^{-1} \cdot \bW) ),
\end{equation}
where $Q$ means a quantization method, $\bLma$ a diagonal transformation matrix, and $\bR$ a orthogonal rotation matrix. $\bW$ and $\bX$ are scaled and rotated to mitigate outliers before implementing quantization, respectively.

%\paragraph{Learnable Singular-value Increment}
\subsection{Learnable Singular-value Increment}
Starting from the singular-value decomposition, LSI first decomposes weight matrix $\bW\in \mathbb{R}^{a\times b}$ (we default that $a\geq b$) into two orthogonal matrices $\mathbf{U}\in \mathbb{R}^{a \times a}$, and $\mathbf{V}\in \mathbb{R}^{b \times b}$ and a matrix $\mathbf{S} \in \mathbb{R}^{b}$ containing non-zero diagonal parameters. Then, it introduces a learnable matrix $\tilde{\mathbf{S}}\in \mathbb{R}^{b}$ that is added on $\mathbf{S}$ during the quantization process:
\begin{equation}
\label{eq:lsi_decompose}
\begin{aligned}
    \tilde{\bW} &= Q(\mathbf{U} \, \mathsf{diag}(\mathbf{S}+\tilde{\mathbf{S}}) \, \mathbf{V}), \\
    \tilde{w}_{ik} &= \sum_{j=0}{u_{ij} (s_{j}+\tilde{s}_{j}) v_{jk}}.
\end{aligned}
\end{equation}
Here, $\mathsf{diag}(\cdot)$ converts the 1-D matrix into a diagonal 2-D one, $\tilde{w}_{ik}$ represents the element of $\tilde{\bW}$ located at the $i$th row and $k$th column. LSI strategically employs a small number of trainable parameters (less than 0.1\% of $\tilde{\bW}$) and achieves both fast training speed and excellent results. 

%------------------------------------------------------
\begin{figure*}[!t]
\includegraphics[width=0.8\textwidth]{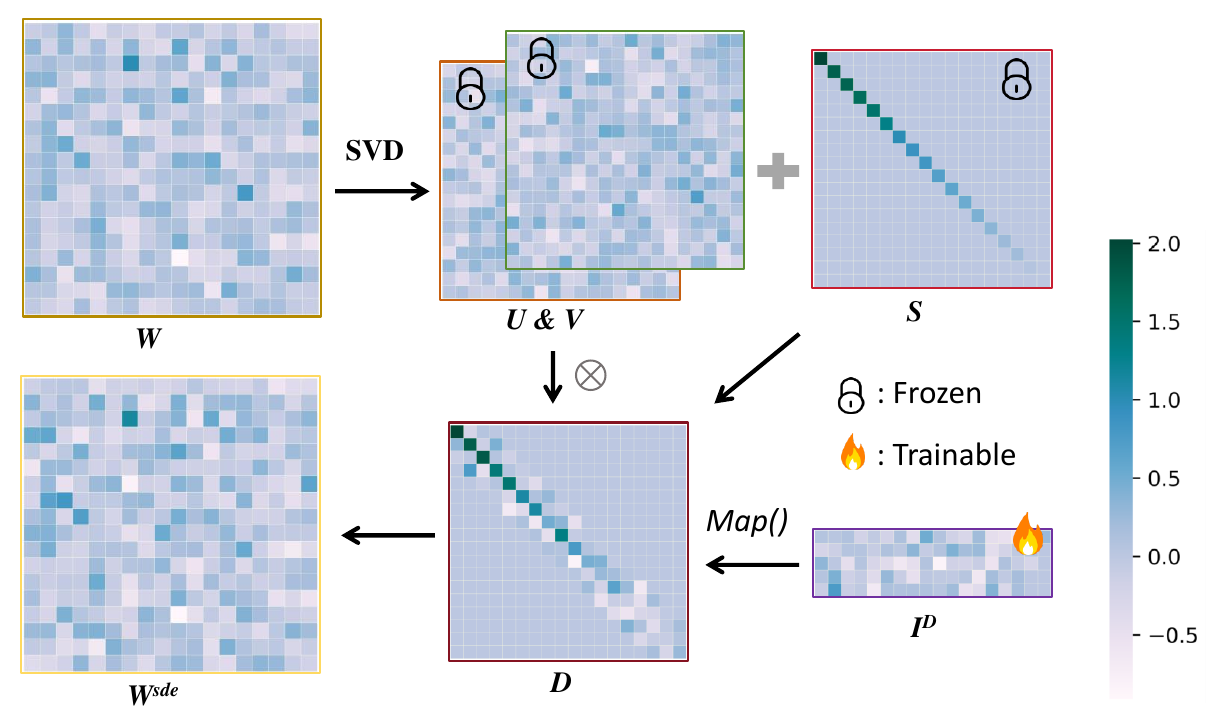}
\centering
%\vspace{-1em}
    \caption{\textbf{Overview of SDE}. After decomposing the linear weight matrix $\mathbf{W}$ into $\mathbf{U}$, $\mathbf{S}$, and $\mathbf{V}$ through singular value decomposition, our technique introduces a learnable matrix $\mathbf{I}^{D}$. This matrix is then appropriately mapped into the diagonal positions of $diag(\mathbf{S})$ using the mapping function $Map(\cdot)$ defined in Eq.~\ref{eq:sde_eq}.}
    \label{fig:sde_ovewiew}
    %\vspace{-1.5em}
\end{figure*}
%------------------------------------------------------

%------------------------------------------------------
\begin{figure}[!t]
\includegraphics[width=0.48\textwidth]{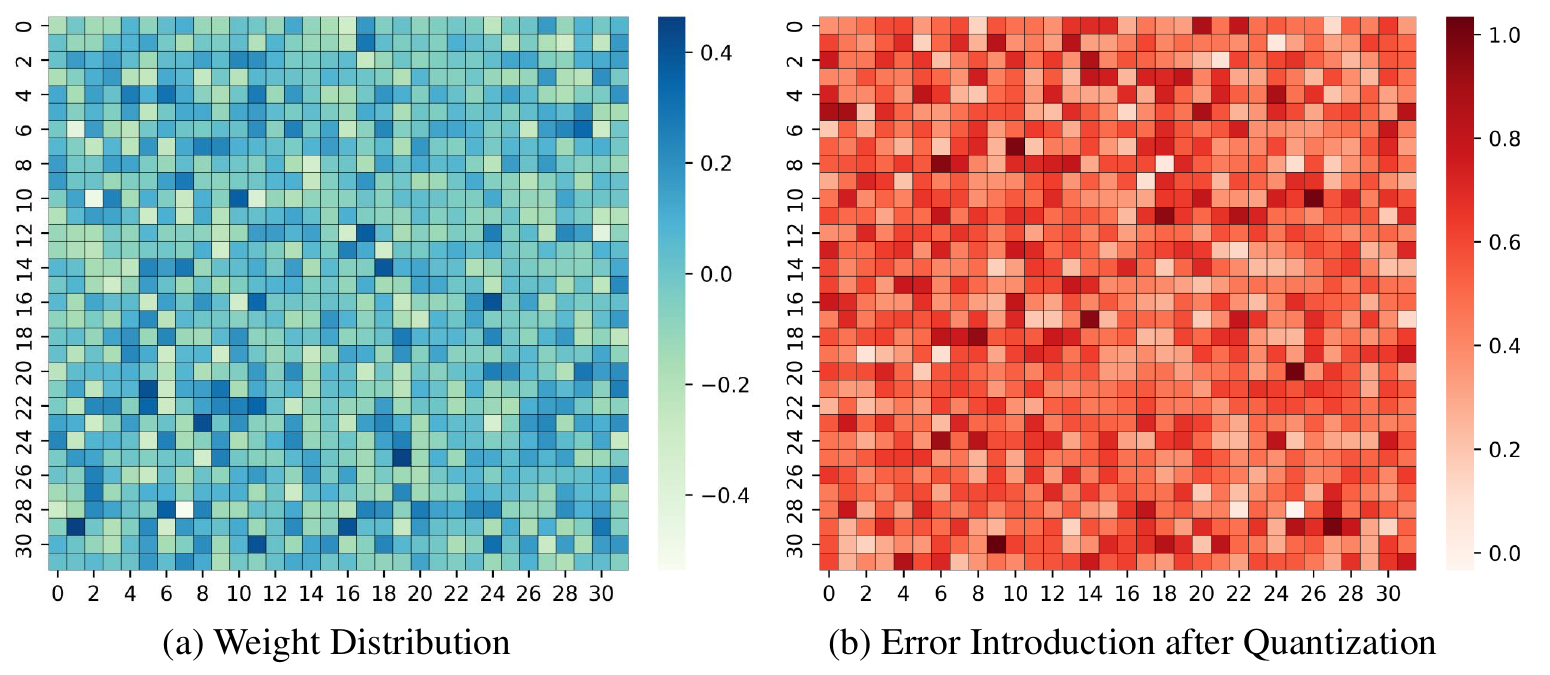}
\centering
%\vspace{-1em}
    \caption{\textbf{Error introduction after quantization.} The weight is from the 2nd layer o-proj in LLaMA-3-8B. The weight matrix is divided into $32 \times 32$ blocks with $128 \times 128$ individual weights in each, and quantization is performed block-by-block. (a) illustrates the mean of normalized weight, while (b) the normalized quantization errors introduced by each block.}
    \label{fig:error_intro}
    %\vspace{-1.5em}
\end{figure}
%------------------------------------------------------

\section{Proposed Method}
This section begins with an analysis of linear weight matrices in models, examining the variance in quantization significance and emphasizing the importance of weight adjustments, which are deficient in LSI. We then present our proposed SDE technique as a targeted solution. Next, we evaluate SDE’s effectiveness mathematically in comparison with LSI. Its overall pipeline of is shown in Fig.~\ref{fig:sde_ovewiew}. Finally, to further improve performance, we introduce CL to redistribute quantization errors across layers. Throughout this paper, all mentions of “weights” refer to the linear weight matrices in LLMs unless stated otherwise.

%------------------------------------------------------
\begin{figure*}[!t]
\includegraphics[width=1\textwidth]{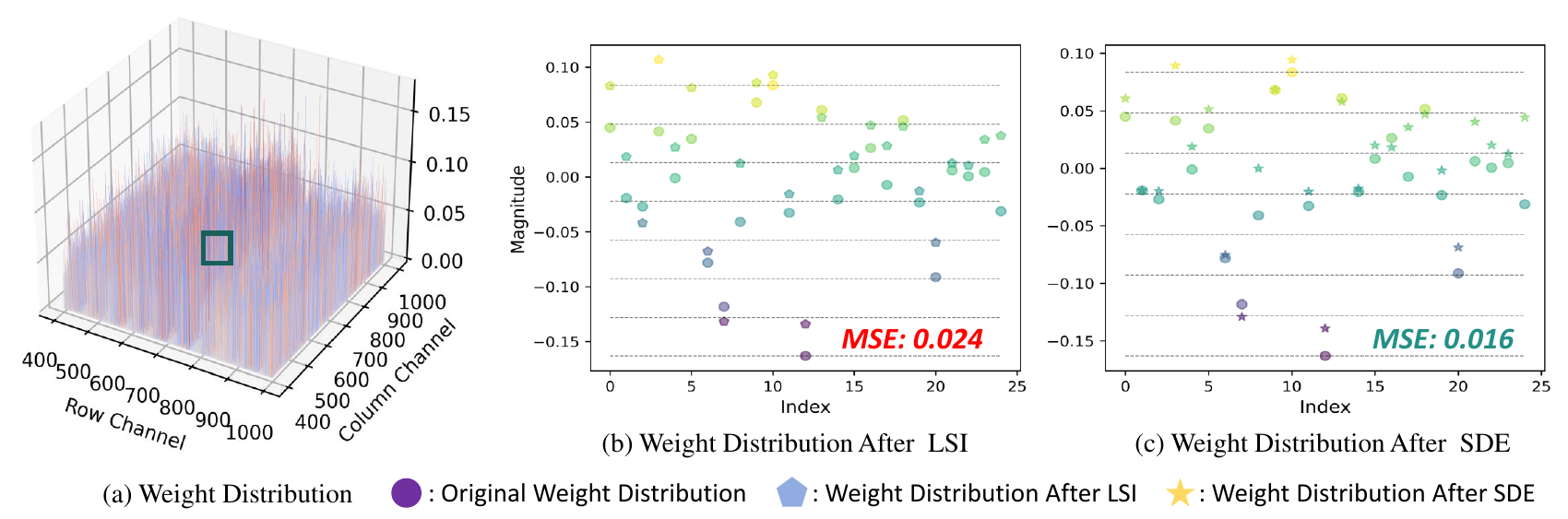}
\centering
%\vspace{-1em}
    \caption{\textbf{Weight redistribution under 3-bit quantization}. The dashed lines represent the corresponding quantized integers after scaling. Both LSI and SDE modify the weight distribution to align the quantization setting, but LSI tends to induce much more disturbance, with MSE compared with original weights a third higher than SDE.}
    \label{fig:weight_disturb}
    %\vspace{-1.5em}
\end{figure*}
%------------------------------------------------------

\subsection{Weight Adjustment Properities}
During quantization, individual weights have varying levels of impact on the final output. A simple verification approach involves making slight modifications to these weights and observing the average error introduced across different outputs, as shown in Figure~\ref{fig:error_intro}. Notably, quantization errors do not consistently correlate with weight magnitude. Moreover, techniques like clipping, scaling, and rotation~\cite{omniquant} add complexity to achieving optimal weight quantization through a fixed approach. Consequently, neither GPTQ nor RTN can address quantization errors at a global level, resulting in non-optimal overall performance. However, GPTQ provides a key improvement over RTN by modifying the Hessian matrix of the weights to facilitate more accurate integer quantization. This suggests that weight-adjustment strategies are conducive to quantization results, as outlined in~\cite{lsi}. Quantization errors can be compensated with each other. These findings help explain why LSI achieves better results to some extent than~\cite{gptq,omniquant,zhao2310atom}.

\subsection{Singular-value Diagonal Expension}
\label{sec:sde}

In our method, we concentrate on the diagonal matrix $\diag(\mathbf{S})$ after SVD and treat its diagonals along different axes as learnable parameters, as illustrated in Fig.~\ref{fig:sde_ovewiew}. Building upon (\ref{eq:lsi_decompose}), the expanded diagonal matrix of learnable singular values is denoted as $\mathbf{I}^{D} \in \mathbb{R}^{(2n+1)\times b}$, where $n$ represents the number of diagonal elements extended along both positive and negative axes. Our method encapsulates the modifications in a formulaic representation:
\begin{equation}
\label{eq:sde_eq}
    \hat{\mathbf{W}} = \mathbf{U} \odot (\diag(\mathbf{S}) + \Map(\mathbf{I}^{D})) \odot \mathbf{V}, \\  
\end{equation}
where $\Map(\cdot)$ is a mapping function that positions the elements of $\mathbf{I}^{D}$ appropriately to multi-axis diagonal elements within $\diag(\mathbf{S})$. Specifically, the first row of $\mathbf{I}^{D}$ with index $0$ is mapped onto the main diagonal. Additionally, the rows at positions $2n-1$ and $2n$ are mapped onto the diagonals of the $n$th axis, which include both the upper and lower axes. For simplicity, we combine $\diag(\mathbf{S})$ and $\Map(\mathbf{I}^{D})$ into a single matrix $\mathbf{D} \in \mathbb{R}^{a\times b}$. Subsequently, for each $d$ in $\mathbf{D}$, the element $\hat{w}_{ik}$ in $\What$ can be derived as follows:
\begin{equation}
\label{eq:sde_ele_eq}
\begin{aligned}
    \hat{w}_{ik} &= \sum_{j=0}^{b} \sum_{k=0}^{b}{u_{ij}d_{jk} v_{kl}},  \\
    \text{where} \: d_{jk} =& 0, \, \text{if} \,\, j < k - n \,\, \text{or} \,\, j > k + n.
\end{aligned}
\end{equation}

%$\lfloor \cdot \rceil$ 

\subsection{Weight Adjustment Flexibility}
\label{sec:weight_adjustment}
From (\ref{eq:quantizer_func}), it is evident that the rounding function grants weight matrix $\mathbf{W}$ the flexibility to have its elements rounded to the same value. This implies that for each element in $\mathbf{W}$, as long as they fall within a specific range, the rounding results are identical. Specifically, if there exists an optimal quantization matrix $\mathbf{W}^Q$ for weight $\mathbf{W}$, our optimization goal is essentially to adjust $\mathbf{W}$’s distribution such that it approximates $\mathbf{W}^Q$ within the permissible rounding error range:
\begin{equation}
\label{eq:opti_goal}
    \lvert \mathbf{W}^Q - \mathbf{W'} \lvert \leq  \mathbf{M}^E , \\
\end{equation}
where $\mathbf{W'}$ is the weight matrix after redistribution, $\mathbf{M}^E$ represents the error matrix, where each element $m^E$ denotes the \textbf{minimum absolute value} of the solution range for each positive or negative side such that $Q(\mathbf{W}') = \mathbf{W}^Q$. As a result, weight quantization becomes an inequality-solving problem. Based on this definition, we could rewrite the optimization goal for LSI in (\ref{eq:lsi_decompose}) as:
\begin{equation}
\label{eq:lsi_opt_ineq}
    \lvert w_{hk}^{Q} - \sum_{j=0}^{b}{u_{ij} (s_{j}+\tilde{s}_{j}) v_{jk}} \lvert  \leq  m_{ik}^E .
\end{equation}
From this inequality, it is clear that every singular-value increment $\tilde{s}_j$ in $\tilde{\mathbf{S}}$ is involved in reducing the error. Although LSI has achieved significant success, solving this inequality remains challenging. This is because the singular values themselves represent less than 1\% of the total weight parameters. Even though singular values are closely tied to the overall weights, altering one singular value can induce changes throughout the entire weight matrix. Through in-depth experiments, we have observed that under a typical MSE loss function, LSI tends to prioritize compensating for large squared errors caused by outliers in a quantization process. This prioritization inadvertently pulls weights with smaller errors into hierarchies that are strongly biased by the current quantization integer hierarchies, as illustrated in Fig.~\ref{fig:weight_disturb}.

On the other hand, our proposed SDV approach retains the global perturbation advantages of LSI while addressing its significant limitations. Based on (\ref{eq:sde_ele_eq}) and (\ref{eq:lsi_opt_ineq}), we reformulate SDE’s optimization objective as:
\begin{equation}
\label{eq:dlsv_ele_ineq}
\begin{aligned}
    \lvert w_{ik}^{Q} -& \sum_{j=0}^{b} \sum_{k=0}^{b}{u_{ij}d_{jk} v_{kl}} \lvert  \leq  m_{ik}^E  \\
    \text{where} \: d_{jk} =& 0, \text{if} \,\, j < k-n \,\, \text{or} \,\, j > k+n. \\
\end{aligned}
\end{equation}
The incorporation of additional learnable parameters through SDE facilitates solving the proposed inequalities in proportion to the value of \( n \). These parameters engage the \(\mathbf{U}\) component more effectively than its usage in LSI, thereby improving the robustness during training and enhancing the generalization during inference.

\subsection{Formal Proof of Greater Flexibility}

In this section, we formally demonstrate that the solution space provided by SDE completely encompasses that of LSI as a special case in Theorem~\ref{thm:sde_superset}:
\begin{theorem}[SDE Strictly Generalizes LSI]
\label{thm:sde_superset}
Let $\Wtilde$ be the set of all weight matrices that can be produced by the LSI parameterization in~\eqref{eq:lsi_decompose}, and let  
$\What$ be the corresponding set produced by the SDE parameterization in~\eqref{eq:sde_ele_eq} with diagonal‑expansion width $n\!\ge\!0$.  
Then, for every $n\ge 0$,
\[
\Wtilde\;\subseteq\;\What_{n}.
\]
When $n=0$ (equivalently, when all off‑diagonal elements added by $\Map(\mathbf{I}^{D})$ are set to zero), the SDE parameterization reduces exactly to LSI.
\end{theorem}

Specifically, let us denote the adjusted singular-value matrix as \( [d_{jl}] \). For the LSI approach described in~\eqref{eq:lsi_decompose}, this adjusted singular-value matrix $[\tilde{d}_{jl}]$ is strictly diagonal:
\begin{equation}
\label{eq:lsi_diag}
[\tilde{d}_{jl}] = (s_j + \tilde{s}_j) \delta_{jl},
\end{equation}
where \(\delta_{jl}\) is the Kronecker delta function (only diagonal values are valid).

In contrast, for the SDE approach defined by~\eqref{eq:sde_ele_eq}, the adjusted singular-value matrix \(\mathbf{D}\) is expanded to include multiple diagonals, as follows:
\begin{equation}
\label{eq:sde_diag}
[\hat{d}_{jl}] =
\begin{cases}
(s_j + i_{0,j}^{D}), & j = l \quad\text{(main diagonal)}\\[6pt]
i_{(n+l-j),j}^{D}, & l-n \leq j \leq l+n, \quad j \neq l \\[6pt]
0, & \text{otherwise.}
\end{cases}
\end{equation}

Clearly, if we set all off-diagonal elements of \(\Map(\mathbf{I}^{D})\) to zero, the parameterization of SDE exactly reduces to that of LSI:
\begin{equation}
\label{eq:sde_inc_lsi}
[\hat{d}_{jl}] = (s_j + i_{0,j}^{D})\delta_{jl} \,\rightarrow\, [\tilde{d}_{jl}] = (s_j + \tilde{s}_j)\delta_{jl}.
\end{equation}
Therefore, SDE’s parameterization inherently includes LSI as a special case.

Consequently, SDE strictly generalizes LSI, implying that the optimal solution space \( \What \) achievable by SDE is a \textbf{superset} of the optimal solution space \( \Wtilde \) achievable by LSI. This expanded flexibility enables SDE to attain equal or lower quantization errors compared to LSI:
\begin{equation}
\label{eq:superset}
\Wtilde  \subseteq \What, \quad \hat{E}_{min} \leq \tilde{E}_{min},
\end{equation}
where \(\tilde{E}_{min}\) and \(\hat{E}_{min}\) represent the minimum achievable errors of LSI and SDE, respectively.

Thus, we conclude that the proposed SDE approach theoretically guarantees superior performance in quantization error minimization over LSI.

% ppl evaluation
\begin{table*}[!t]
\footnotesize
\caption{Perplexity ($\downarrow$) results under 4-bit weight-activation quantization for LLaMA families on WikiText. For OmniQuant, both 3-8B and 3-70B models utilize lwc exclusively during quantization. Our techniques consistently enable various baselines to gain further performance improvements.
}
\vspace{-10pt}
\begin{center}
% \resizebox{0.82\linewidth}{!}{
%\resizebox{1\linewidth}{!}
{
\begin{tabular}{l|ccccccccc}

\toprule
\multicolumn{1}{l|}{\textbf{Method}} & \textbf{1-7B} & \textbf{1-13B} & \textbf{1-30B} & \textbf{1-65B} & \textbf{2-7B} & \textbf{2-13B} & \textbf{2-70B} & \textbf{3-8B} & \textbf{3-70B} \\ 
\midrule
  FP16                            & 5.68          & 5.09           & 4.10           & 3.53           & 5.47          & 4.88           & 3.31       & 6.14         &  2.58  \\ 
\midrule
    QLLM  \cite{liu2023qllm}       & 9.65          & 8.41           & 8.37           & 6.87           & 11.75         & 9.09           & 7.00          & -       &  -      \\
    Atom   \cite{zhao2310atom}     & 8.15          & 7.43           & 6.52           & 5.14           & 8.40          & 6.96           & NaN           & -    & -       \\
    QuaRot \cite{ashkboos2024quarot}  & -   & -  & -   & -   & 6.19  & 5.45   & 3.83   & 8.41    & 6.82      \\
\midrule
       OmniQuant  \cite{omniquant}              & 11.26         & 10.87          & 10.33          & 9.17           & 14.26         & 12.30          & NaN          & 4.5e+3      & NaN         \\
       OmniQuant-LSI  \cite{lsi}                    & 11.02         & 10.68          & 10.20          & 8.95           & 13.78         & 12.04          & 456          & 337      & 516         \\
       \cellcolor{purple!10}\textbf{Ours}-OmniQuant    & \cellcolor{purple!10}10.21         & \cellcolor{purple!10}9.72          & \cellcolor{purple!10}9.12          & \cellcolor{purple!10}8.33           & \cellcolor{purple!10}12.74         & \cellcolor{purple!10}10.96          & \cellcolor{purple!10}21.57          & \cellcolor{purple!10}32.64      & \cellcolor{purple!10}61.85        \\
\midrule
    DuQuant \cite{lin2024duquant} & 6.40  & 5.65   & 4.72  & 4.13   & 6.28  & 5.42   & 3.79  & 8.14   & 5.67 \\
    \cellcolor{purple!10}\textbf{Ours}-DuQuant    & \cellcolor{purple!10}6.26         & \cellcolor{purple!10}5.55          & \cellcolor{purple!10}4.66          & \cellcolor{purple!10}4.02           & \cellcolor{purple!10}6.09         & \cellcolor{purple!10}5.34          & \cellcolor{purple!10}3.71          & \cellcolor{purple!10}7.93     & \cellcolor{purple!10}5.55      \\
\hline
    DuQuant FT           & 6.18 & 5.47  & 4.55  & 3.93  & 6.08 & 5.33  & 3.76  & 7.84  & 5.59\\ 
    \cellcolor{purple!10}\textbf{Ours}-DuQuant FT    & \cellcolor{purple!10}6.07     & \cellcolor{purple!10}\textbf{5.41}   & \cellcolor{purple!10}\textbf{4.46}    & \cellcolor{purple!10}\textbf{3.85}    & \cellcolor{purple!10}5.99         & \cellcolor{purple!10}5.26         & \cellcolor{purple!10}\textbf{3.68}          & \cellcolor{purple!10}7.67      & \cellcolor{purple!10}5.51        \\
\midrule
    PrefixQuant w/o FT & 6.34   & 5.82  & 4.83    & 4.22   & 6.22  & 5.50  & 4.41  & 7.93 & 5.23 \\
    
    \cellcolor{purple!10}\textbf{Ours}-Prefix w/o FT    & \cellcolor{purple!10}6.13        & \cellcolor{purple!10}5.71          & \cellcolor{purple!10}4.76          & \cellcolor{purple!10}4.16           & \cellcolor{purple!10}6.06         & \cellcolor{purple!10}5.29          & \cellcolor{purple!10}4.27          & \cellcolor{purple!10}7.64      & \cellcolor{purple!10}5.11        \\

\hline
    PrefixQuant FT & 6.12   & 5.68  & 4.72   & 4.09   & 6.01  & 5.32  & 3.81  & 7.43 & 4.41 \\
    
    \cellcolor{purple!10}\textbf{Ours}-Prefix FT    & \cellcolor{purple!10}\textbf{5.96}         & \cellcolor{purple!10}5.61          & \cellcolor{purple!10}4.59          & \cellcolor{purple!10}3.97           & \cellcolor{purple!10}\textbf{5.92}         & \cellcolor{purple!10}\textbf{5.15}          & \cellcolor{purple!10}3.69          & \cellcolor{purple!10}\textbf{7.24}      & \cellcolor{purple!10}\textbf{4.29}       \\
\bottomrule
\end{tabular}

}
\end{center}
\vspace{-3.5mm}
\label{tab:ppl-w4a4-wiki}
\end{table*}

\subsection{Cross-layer Learning}
\label{sec:cl}
Existing PTQ methods often focuses on maximizing convenience, limiting learning-based optimization to individual layers. As a result, under certain conditions, they produce inferior results compared to QAT. Conversely, QAT emphasizes the consistency of the overall output of the model, employing global training of model weights to achieve superior results—albeit at a significantly higher computing cost. To enhance inter-layer connectivity while maintaining the lightweight nature of PTQ, we propose Cross-layer Learning to train the model. During the few-sample learning process of the current layer, we not only generate the quantized output for that layer, but also propagate it through the subsequent layer, capturing the resulting output. Thus, our loss is influenced not only by the current layer’s pre- and post-quantization mean-square error (MSE) but also by the MSE between the next layer’s output and the output produced by feeding it the current layer’s quantized input. 

In the training process, the parameters of the subsequent layer do not undergo any additional training. They only allow gradient propagation. For the $i$th layer's output $\mathbf{Y}^i$ and quantized output $\mathbf{Y}^{Q^{i}}$, our loss is computed as:
\begin{equation}
\label{eq:final_loss}
    \mathcal{L} = \mathcal{L}_2(\mathbf{Y}^i - \mathbf{Y}^{Q^{i}}) + \lambda \mathcal{L}_2(\mathbf{Y}^{i+1} - \mathbf{Y}^{Q^{i+1}}),
\end{equation}
where $\mathcal{L}_2$ means MSE loss, $\lambda$ the loss scaler, $\mathbf{Y}$ and $\mathbf{Y}^{Q}$ are the original outputs and its quantized outputs, respectively. This approach avoids significant increases in training resource consumption while maintaining inter-layer connections. As a result, quantization errors are more evenly distributed across the model’s layers, achieving superior results over existing PTQ methods.

\section{Experiments}
%\midrule[0.7pt]
%\bottomrule[1pt]

%\input{latex/tables/l-w4a4-ppl-c4}

\subsection{Settings}

\paragraph{Baselines and Implementation}
Our experiments cover both weight-only and weight-activation quantization scenarios. For weight-activation and KV Cache quantization, we select previously state-of-the-art baselines including SmoothQuant~\cite{smoothquant}, Outlier Suppression+~\cite{outlier-plus}, OmniQuant~\cite{omniquant}, QLLM~\cite{liu2023qllm}, Atom~\cite{zhao2310atom}, QuaRot~\cite{ashkboos2024quarot}, DuQuant~\cite{lin2024duquant}, and PrefixQuant~\cite{chen2024prefixquant}. Since our approach is plug-and-play, we integrate our techniques into three baselines—OmniQuant, DuQuant, and PrefixQuant—to demonstrate both the effectiveness and generality of our methods. It is important to note that for OmniQuant, DuQuant, and PrefixQuant, we do not alter any of their original experimental configurations, except for the incorporation of SDE and CL. Specifically, all DuQuant experiments are conducted with the smoothing technique~\cite{smoothquant}, consistent with the original implementation. FT for DuQuants indicates the use of LWC~\cite{omniquant} for fine-tuning, whereas for PreFixQuant, fine-tuning is used the same as that in the original paper. For downstream tasks and open-source benchmarks, DuQuant may not necessarily achieve the best results after fine-tuning, but our techniques always require calibration, and so we select the best performance among both DuQuant and DuQuant-FT, and indicate it as DuQuant-Best. For weight-only quantization, we compare against such pure weight quantization methods as GPTQ~\cite{gptq}, LSI~\cite{lsi}, and RTN to demonstrate the advantages of our methods.

\paragraph{Training Configurations} 
During training, we set our SDE parameters with a low learning rate of $2.5e-4$, matching that of LSI, to minimize disturbances. For CL, we set $\lambda=0.05$ in~\eqref{eq:final_loss} to avoid next-layer loss confusion. The training data consists of 128 samples, each with a sequence length of 2048, sourced from WikiText2~\citep{wikitext2}. The expanded diagonal number $n$ is consistently set to 100 throughout the experiments.

\paragraph{Models} We conduct experiments on contemporary popular baselines, LLaMA (1-3)~\citep{llama} and OPT~\citep{opt} for basic weight-only and weight-activation quantization settings, and instruction-tuned LLMs: Vicuna-v1.5 (7B-13B)~\cite{vicuna} for open-source benchmarks.

\paragraph{Evaluation} Our evaluation of perplexity primarily focuses on the WikiText~\citep{wikitext2} and C4~\citep{c4} datasets. In line with prior research for zero-shot inference, we assess 5 common-sense tasks including PIQA~\citep{piqa}, ARC (ARC-E and ARC-C)~\citep{arc}, HellaSwag (HS)~\citep{hellaswag}, Winogrande (WG)~\cite{winogrande} and Boolq~\cite{boolq}. Moreover, to highlight the advantages of our method over LSI, we additionally include evaluations on the QQP and MRPC datasets in GLUE~\cite{glue}. We use the LM
Evaluation Harness~\cite{gao2021lmharness} with default parameters for our experiments. Moreover, we test the open-source benchmark MT-Bench~\cite{zheng2023mtbench} under GPT-4 evaluation protocol~\cite{vicuna} and LongBench~\cite{bai2023longbench} for accuracy testing. We use GPT-4o as the evaluation model.

%----------------------------------------------
%We report \textit{acc} for WinoGrande and \textit{acc norm} for HellaSwag, Arc-C, Arc-E, and PIQA, following Qserve~\cite{lin2024qserve}.

%GLUE (RTE, QQP, MRPC)~\cite{glue}, WIC~\cite{superglue}, COQA~\cite{coqa}, Winogrande (WG)~\cite{winogrande} and Boolq~\cite{boolq}. The datasets selected for evaluation adhere to the GPTQ~\citep{gptq} guidelines. For measuring accuracy in these zero-shot tasks, we utilize the lm-eval-harness~\citep{evaluation}.

\subsection{Quantization Results}
\paragraph{Weight-activation Quantization Results}
From Tables~\ref{tab:ppl-w4a4-wiki}-\ref{tab:llama-w3a4}, it is evident that our method consistently outperforms various baselines, achieving significant improvements in both perplexity (PPL) and downstream task performance, while maintaining the best overall results. Starting with the common W4A4 setting, for OmniQuant on the LLaMA-3 series, even in cases where the original method fails entirely and results in model inference collapse, our approach manages to keep PPL within a range that allowes basic inference functionality. LSI also achieves a certain level of improvement; however, the magnitude of this improvement diminishes significantly as the model size increases. It fails to produce usable results for LLaMA-3 families and LLaMA-2-70B. Besides, performance on certain downstream tasks using LSI can degrade substantially, to be illustrated later.

For DuQuant and PrefixQuant, our method reduces the average PPL by over 0.1. In terms of common task performance, our method improves OmniQuant by more than 1.5\% on average and DuQuant by more than 1\%. Even for PrefixQuant’s nearly lossless quantization results, we achieves a measurable improvement. In the W3A4 setting, the improvements are more pronounced. Compared to the W4A4 setting, whereas the original DuQuant nearly doubles PPL and significantly reduces downstream task performance, our approach remains robust, maintaining much of the performance.

% ppl evaluation
\begin{table*}[!t]
\footnotesize
\caption{Downstream tasks' performance ($\uparrow$) under W4A4KV4 for LLaMA families. We report average performance for each model on 5 tasks. For OmniQuant, our calculation does not include 2-70B. DuQuant(Best) means we take the highest performance of DuQuant and DuQuant-FT.}
\vspace{-10pt}
\begin{center}
% \resizebox{0.82\linewidth}{!}{
%\resizebox{0.92\linewidth}{!}
{
\begin{tabular}{l|ccccccc|c}
\toprule
\multicolumn{1}{l|}{\textbf{Method}} & \textbf{1-7B} & \textbf{1-13B} & \textbf{1-30B} & \textbf{1-65B} & \textbf{2-7B} & \textbf{2-13B} & \textbf{2-70B}& \textbf{Average}\\ 
\midrule
  FP16  & 64.09  & 66.33   & 67.44   & 71.04  & 63.72   & 66.08   & 70.39 & 67.01 \\ 
\midrule
    QLLM  \cite{liu2023qllm}  & 51.84  & 55.28  & 57.87   & 59.83  & 51.60    & 54.31  & 58.62  & 55.62    \\
    SmoothQuant  \cite{smoothquant}  & 38.41  & 49.36  & 44.83   & 47.71  & 44.52    & 47.67  & 50.30  & 46.11    \\
    Outlier Sup+  \cite{outlier-plus}  & 48.43  & 49.86  & 52.62   & 52.52  & 45.66    & 49.76  & 50.73  & 49.94    \\
    AffineQuant  \cite{ma2024affinequant}  & 53.42  & 52.58  & 58.61   & -  & 52.64    & 55.09  & -  & 54.46    \\
    Atom  \cite{zhao2310atom}  & 56.88  & 58.04  & 59.50   & 59.83  & 55.58    & 58.29  & -  & 58.02    \\
\midrule
       OmniQuant  \cite{omniquant}  & 52.65    & 54.37   & 56.63   & 59.22  & 51.38   & 55.19  & -   & 54.90       \\
       %OmniQuant-LSI    & 53.82    & 54.91   & 57.02   & 59.38  & 53.04   & 55.65  & -   & 55.63       \\
       \cellcolor{purple!10}\textbf{Ours}-OmniQuant    & \cellcolor{purple!10}54.47         & \cellcolor{purple!10}55.86          & \cellcolor{purple!10}57.91          & \cellcolor{purple!10}60.09           & \cellcolor{purple!10}53.15         & \cellcolor{purple!10}56.77          & \cellcolor{purple!10}-      & \cellcolor{purple!10}56.37(\textbf{+1.47})     \\
\midrule
    DuQuant-Best \cite{lin2024duquant} & 61.76  & 64.12  & 65.15 & 68.91  & 60.57  & 63.47   & 67.89  & 64.55 \\
    \cellcolor{purple!10}\textbf{Ours}-DuQuant    & \cellcolor{purple!10}\textbf{62.29}  & \cellcolor{purple!10}64.01          & \cellcolor{purple!10}\textbf{66.23}  & \cellcolor{purple!10}\textbf{69.15}  & \cellcolor{purple!10}61.17         & \cellcolor{purple!10}\textbf{63.95}  & \cellcolor{purple!10}68.34          & \cellcolor{purple!10}\textbf{65.02}(+0.47)    \\
\midrule
    PrefixQuant FT & 61.14   & 63.57  & 65.45    & 68.21   & 61.58  & 63.03  & 68.11  & 64.44\\
    
    \cellcolor{purple!10}\textbf{Ours}-Prefix FT    & \cellcolor{purple!10}61.56        & \cellcolor{purple!10}63.83          & \cellcolor{purple!10}65.76          & \cellcolor{purple!10}68.48           & \cellcolor{purple!10}\textbf{62.14}   & \cellcolor{purple!10}63.47          & \cellcolor{purple!10}68.52         & \cellcolor{purple!10}64.82(+0.38)    \\
\bottomrule
\end{tabular}

}
\end{center}
\vspace{-3.5mm}
\label{tab:llama-perf-w4a4}
\end{table*}

%------------------------------------------------------------
\begin{table}[tb]
\centering
\footnotesize
%\normalsize
\caption{W3A4 quantization results of LLaMA models. Average refers to average performance on 5 downstream tasks.}
\label{tab:llama-w3a4}
%\resizebox{\linewidth}{!}
{
\begin{tabular}{l|l|cc|c}
\toprule
\multicolumn{2}{l|}{\textbf{Model \& Method}} & WIKI & C4  & Average \\ 
\midrule

\multirow{2}{*}{1-7B} 
& DuQuant-FT & 11.70  &  13.55  & 50.49 \\
&\cellcolor{purple!10}\textbf{Ours} 
&\cellcolor{purple!10}\textbf{7.31} 
&\cellcolor{purple!10}\textbf{9.33} 
&\cellcolor{purple!10}\textbf{56.94} \\ 
\midrule

\multirow{2}{*}{1-13B} 
& DuQuant-FT & 8.21  &  11.04  & 52.79 \\
&\cellcolor{purple!10}\textbf{Ours} 
&\cellcolor{purple!10}\textbf{6.11}  
&\cellcolor{purple!10}\textbf{7.49} 
&\cellcolor{purple!10}\textbf{57.94} \\  
\midrule

\multirow{2}{*}{2-7B} 
& DuQuant-FT & 15.29  &  17.58  & 49.06 \\
&\cellcolor{purple!10}\textbf{Ours} 
&\cellcolor{purple!10}\textbf{7.50}  
&\cellcolor{purple!10}\textbf{9.78} 
&\cellcolor{purple!10}\textbf{56.45} \\  
\midrule

\multirow{2}{*}{2-13B} 
& DuQuant-FT & 8.34  &  11.75  & 52.33 \\
&\cellcolor{purple!10}\textbf{Ours} 
&\cellcolor{purple!10}\textbf{6.35}  
&\cellcolor{purple!10}\textbf{7.96} 
&\cellcolor{purple!10}\textbf{59.12} \\ 

\bottomrule
\end{tabular}
}
\end{table}
%------------------------------------------------------------

\begin{table*}[!t]
\caption{
Zero-shot and five-shot results on the MMLU benchmark for Vicuna-v1.5-13B under 4-bit weight-activation quantization.
}
\vspace{-10pt}
\begin{center}
\resizebox{0.95\linewidth}{!}{
\begin{tabular}{c|l|ccccc|ccccc}
\toprule
\multirow{2}{*}{\textbf{Model}}  & \multirow{2}{*}{\textbf{Method}} & \multicolumn{5}{c|}{\textbf{MMLU (0 shot) $\uparrow$}}          & \multicolumn{5}{c}{\textbf{MMLU (5 shot) $\uparrow$}}  \\ \cmidrule{3-12} 
                                 &                                  
&\textbf{STEM}  & \textbf{Hums}  & \textbf{Social} & \textbf{Others} & \textbf{Avg.}  & \textbf{STEM}   & \textbf{Hums}  & \textbf{Social} & \textbf{Others} & \textbf{Avg.}  \\ 
\midrule
\multirow{7}{*}{\begin{tabular}[c]{@{}c@{}}Vicuna-v1.5-13B\\ W4A4\end{tabular}}
& FP16                    & 43.70          & 50.48          & 62.72          & 62.74          & 54.54          & 44.96           & 51.97          & 65.26          & 62.40                              & 55.78          \\
\cmidrule(lr){2-12}
& SmoothQuant  \cite{smoothquant}     & 21.70          & 24.29          & 22.13          & 23.16          & 22.82          & 25.31           & 24.97          & 26.00          & 27.08                              & 25.84          \\
& Atom    \cite{zhao2310atom}       & 32.54          & 39.60          & 46.02           & 46.11         & 41.07          & 35.35           & 39.21          & 59.72           & 45.77           
           & 45.01          \\
\cmidrule(lr){2-12}
& OmniQuant  \cite{omniquant}   & 26.81          & 26.57          & 30.35          & 28.75          & 28.12          & 28.79           & 27.29          & 31.13          & 28.99                              & 29.05          \\
           
&  \cellcolor{purple!10}\textbf{Ours}-OmniQuant & \cellcolor{purple!10}39.06 & \cellcolor{purple!10}35.68          & \cellcolor{purple!10}41.74          & \cellcolor{purple!10}37.69          & \cellcolor{purple!10}38.54          & \cellcolor{purple!10}34.16          & \cellcolor{purple!10}37.57 & \cellcolor{purple!10}45.23 & \cellcolor{purple!10}37.96                        & \cellcolor{purple!10}38.73 \\

\cmidrule(lr){2-12}
& DuQuant-Best  \cite{lin2024duquant}   & 40.82          & 47.48          & 58.86          & 57.83          & 51.24          & 41.42           & 48.78          & 60.42          & 57.74                              & 52.09          \\

&  \cellcolor{purple!10}\textbf{Ours}-DuQuant   & \cellcolor{purple!10}\textbf{41.79}          & \cellcolor{purple!10}\textbf{48.56} & \cellcolor{purple!10}\textbf{60.12} & \cellcolor{purple!10}\textbf{58.47} & \cellcolor{purple!10}\textbf{52.23} &  \cellcolor{purple!10}\textbf{42.16} & \cellcolor{purple!10}\textbf{49.69} & \cellcolor{purple!10}\textbf{62.74} & \cellcolor{purple!10}\textbf{60.03} & \cellcolor{purple!10}\textbf{53.65}  \\ 
\bottomrule
\end{tabular}
}
\end{center}
\vspace{-1.em}
\label{tab:mmlu}
\end{table*}

% longcontext
\begin{table*}[!t]
\caption{
Long-context generation results for 4-bit Vicuna models on the LongBench benchmark.
}
\vspace{-10pt}
\begin{center}
\resizebox{0.95\linewidth}{!}{
\begin{tabular}{c|l|ccccccccc}
\toprule
\textbf{Vicuna}                                                                 & \multicolumn{1}{c|}{\textbf{Setting}} & \textbf{Qasper} & \textbf{QMSum} & \textbf{MultiNews} & \textbf{TREC}  & \textbf{TriviaQA} & \textbf{SAMSum} & \textbf{DuReader} & \textbf{RepoBench-P} & \textbf{Avg}   \\ \midrule
\multirow{7}{*}{\begin{tabular}[c]{@{}c@{}}Vicuna-v1.5-7B\\ W4A4\end{tabular}}  
& FP16                                  & 23.27           & 21.07          & 26.91              & 66.00          & 82.59             & 41.06           & 25.53             & 48.23                & 41.83          \\
\cmidrule(lr){2-11}
& SmoothQuant   \cite{smoothquant}    & 4.11            & 2.00           & 6.05               & 15.00          & 1.62              & 1.55            & 4.24              & 25.92                & 7.56           \\
& Atom   \cite{zhao2310atom}                & 17.97           & 20.24          & 24.60              & 58.00          & 67.20             & 37.94           & 19.41             & 29.34                & 34.34          \\

\cmidrule(lr){2-11}

& OmniQuant  \cite{omniquant}  & 1.62            & 3.93           & 2.64               & 1.00           & 0.81              & 0.61            & 1.87              & 14.97                & 3.43           \\

& \cellcolor{purple!10}\textbf{Ours}-OmniQuant              & \cellcolor{purple!10}12.14  & \cellcolor{purple!10}11.77 & \cellcolor{purple!10}15.05     & \cellcolor{purple!10}32.47 & \cellcolor{purple!10}41.25    & \cellcolor{purple!10}22.05  & \cellcolor{purple!10}13.77    & \cellcolor{purple!10}17.35       & \cellcolor{purple!10}20.73 \\ 

\cmidrule(lr){2-11}

& DuQuant-Best     & 19.98            & \textbf{21.15}           & 25.85               & 64.00           & 78.91              & \textbf{42.24}            & 23.15              & 47.66                & 40.37           \\

& \cellcolor{purple!10}\textbf{Ours}-DuQuant              & \cellcolor{purple!10}\textbf{20.35}  & \cellcolor{purple!10}20.96 & \cellcolor{purple!10}\textbf{26.34}     & \cellcolor{purple!10}\textbf{64.48} & \cellcolor{purple!10}\textbf{80.16}    & \cellcolor{purple!10}41.65  & \cellcolor{purple!10}\textbf{24.48}    & \cellcolor{purple!10}\textbf{47.99}       & \cellcolor{purple!10}\textbf{40.81} \\

\midrule
\multirow{7}{*}{\begin{tabular}[c]{@{}c@{}}Vicuna-v1.5-13B\\ W4A4\end{tabular}} 
& FP16                                  & 24.41           & 21.24          & 26.53              & 68.00          & 86.81             & 41.97           & 27.57             & 43.08                & 42.45          \\
\cmidrule(lr){2-11}
& SmoothQuant   \cite{smoothquant}        & 2.18            & 2.95           & 3.54               & 1.50           & 1.83              & 0.35            & 6.71              & 11.57                & 3.83           \\
& Atom   \cite{zhao2310atom}   & 17.67           & 20.23          & 23.39              & 59.00          & 80.75             & 38.72           & 21.79             & 37.31                & 37.36          \\
\cmidrule(lr){2-11}
& OmniQuant    \cite{omniquant}        & 0.68            & 1.78           & 2.83               & 9.00           & 1.13              & 0.45            & 13.83             & 8.46                 & 4.77           \\
& \cellcolor{purple!10}\textbf{Ours}-OmniQuant              & \cellcolor{purple!10}11.47  & \cellcolor{purple!10}13.52 & \cellcolor{purple!10}16.93     & \cellcolor{purple!10}45.36 & \cellcolor{purple!10}51.59    & \cellcolor{purple!10}23.14  & \cellcolor{purple!10}15.69    & \cellcolor{purple!10}21.76       & \cellcolor{purple!10}24.93 \\ 

\cmidrule(lr){2-11}
& DuQuant-Best \cite{lin2024duquant} & 18.93            & 20.72      & \textbf{26.59}               & \textbf{66.50}           & \textbf{83.04}    & 42.67            & \textbf{26.02}             & 38.09     & 40.32          \\
& \cellcolor{purple!10}\textbf{Ours}-DuQuant                & \cellcolor{purple!10}\textbf{20.85}  & \cellcolor{purple!10}\textbf{20.97} & \cellcolor{purple!10}26.14     & \cellcolor{purple!10}\textbf{66.50} & \cellcolor{purple!10}82.15    & \cellcolor{purple!10}\textbf{42.89}  & \cellcolor{purple!10}25.16    & \cellcolor{purple!10}\textbf{45.83}      & \cellcolor{purple!10}\textbf{41.31} \\ 

\bottomrule
\end{tabular}
}
\end{center}
\vspace{-1.em}
\label{tab:longbench}
\end{table*}

\paragraph{Weight-only Quantization Results}

In this section, we showcase the quantization results of the OPT series on PPL, as detailed in Table~\ref{tab:opt-w-only}. Compared to GPTQ and LSI, our approach consistently achieves the best quantization results without introducing the significant sample bias associated with LSI. The advantages of our approach become increasingly evident as the quantization bit size decreases. In the case of larger LLMs, such as the 66B model at 2 bits, where both GPTQ and LSI fail to achieve meaningful quantization results, our method successfully quantizes the model to a level that supports normal inference. All of these advancements strongly highlight the superiority of our approach over the existing ones.

%------------------------------------------------------
\begin{figure}[!t]
\includegraphics[width=0.48\textwidth]{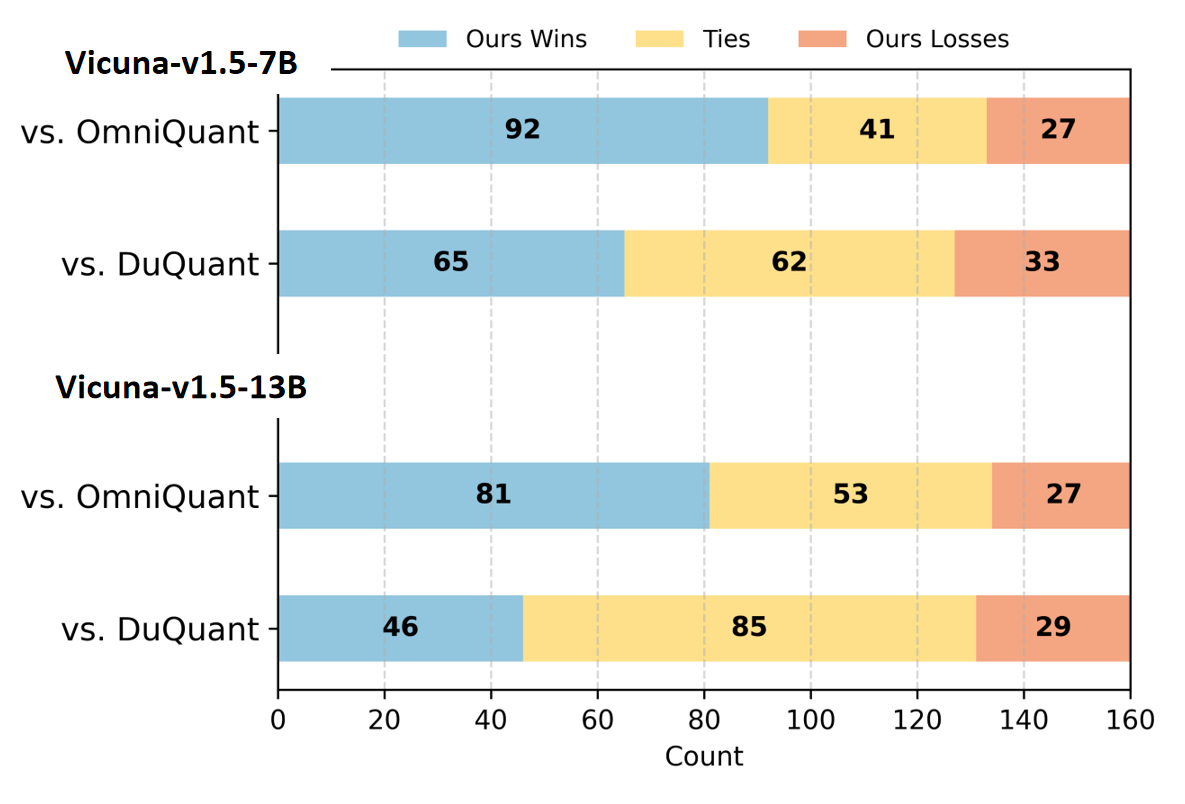}
\centering
%\vspace{-1em}
    \caption{GPT-4 evaluation on the MT-Bench.}
    \label{fig:gpt4_eval}
    %\vspace{-1.5em}
\end{figure}
%--------

\paragraph{Open-source Benchmark Evaluation}

Our method consistently improves performance across a variety of baselines and benchmarks. As shown in Fig.~\ref{fig:gpt4_eval}, on MT-Bench, LLMs quantized using our approach demonstrate significantly better performance. In particular, for OmniQuant, our results substantially surpass those of the original methods. For DuQuant, given its more mature quantization mechanism, the performance gains are especially prominent on the 7B models, with winning 65 cases.

Furthermore, as shown in Tabs.~\ref{tab:mmlu} and \ref{tab:longbench} on MMLU and LongBench, our technique provides especially notable improvements for OmniQuant on complex tasks, with gains reaching up to 10\%, strongly validating the effectiveness of our method. For DuQuant, our method better preserves the model’s original generation capabilities. In cases where the original baseline performs abnormally (accuracy higher than the full-precision model), we offer a degree of correction and stabilization. Conversely, for tasks where performance drops significantly below that of the original model, our method better retains the inherent characteristics of the model, as exemplified by the results on Vicuna-v1.5-13B for RepoBench-P in Tab.~\ref{tab:longbench}.

%---------------------------------------------------------------------
\begin{table*}[t]
%\normalsize
\footnotesize
    \setlength{\tabcolsep}{6pt}
    \centering
    \caption{Perplexity of weight-only quantization in OPT models. All methods use WikiText for calibration.} 
    %\begin{threeparttable}
    \begin{tabular}{l|l|cc|cc|cc|cc|cc|cc}
        \toprule
        \multicolumn{2}{l|}{\textbf{Methods / PPL} $\downarrow$}  & \multicolumn{2}{c|}{1.3B} & \multicolumn{2}{|c|}{2.7B} & \multicolumn{2}{c|}{6.7B} & \multicolumn{2}{c|}{13B} & \multicolumn{2}{c|}{30B} & \multicolumn{2}{c}{66B}\\ 
        \midrule
        \multicolumn{2}{c|}{-} & Wiki & C4 & Wiki & C4 & Wiki & C4 & Wiki & C4 & Wiki & C4 & Wiki & C4 \\ 
        \midrule
        FP16 & - & 14.63 & 10.86 & 12.47 & 9.56 & 10.86  & 14.63 & 10.12 & 10.86 & 9.56 & 9.56 & 9.34 & 9.34\\ 
    \midrule

        \multirow{4}{*}{\shortstack{W2A16\\g128}} & RTN & 1.3e4 & 7.7e3 & 5.7e4 & 3.8e4 & 7.8e3  & 5.2e3 & 7.6e4 & 2.8e4 & 1.3e4 & 6.5e3 & 3.6e5 & 2.6e5 \\ 
        & GPTQ \cite{gptq} & 115 & 60.88 & 61.59 & 33.83 & 20.18  & 18.55 & 21.36 & 16.34 & 12.71 & 12.89 & 82.10 & 598.81\\ 
        & LSI \cite{lsi} & 2.8e3 & 2.4e3 & 379 & 296 & 46.15  & 39.88 & 24.92 & 23.57 & 12.64 & 13.29 & 408 & 492\\
        %\cmidrule{2-14}
        &\cellcolor{Gray}\textbf{Ours} 
        &\cellcolor{Gray}\textbf{30.54}
        &\cellcolor{Gray}\textbf{32.77}
        &\cellcolor{Gray}\textbf{24.66}
        &\cellcolor{Gray}\textbf{26.94} 
        &\cellcolor{Gray}\textbf{18.36}  
        &\cellcolor{Gray}\textbf{17.14} 
        &\cellcolor{Gray}\textbf{16.55} 
        &\cellcolor{Gray}\textbf{15.27} 
        &\cellcolor{Gray}\textbf{12.04} 
        &\cellcolor{Gray}\textbf{12.21} 
        &\cellcolor{Gray}\textbf{15.88} 
        &\cellcolor{Gray}\textbf{16.51}\\ 

    \midrule

        \multirow{4}{*}{\shortstack{W2A16\\g64}} & RTN & 1.0e4 & 7.3e3 & 19.3e4 & 1.2e5 & 7.6e3  & 6.3e3 & 1.8e4 & 7.5e3 & 8.2e3 & 4.0e3 & 1.1e4 & 8.4e3 \\ 
        & GPTQ \cite{gptq} & 49.58 & 31.31 & 29.37 & 23.23 & 16.81  & 16.24 & 16.65 & 14.48 & 11.87 & 12.24 & 356.01 & 58.60\\ 
        & LSI \cite{lsi} & 32.11 & 34.53 & 24.18 & 25.44 & 15.79  & 16.88 & 15.08 & 15.33 & 12.35 & 13.11 & 218 & 337\\
        %\cmidrule{2-14}
        &\cellcolor{Gray}\textbf{Ours} 
        &\cellcolor{Gray}\textbf{28.08}
        &\cellcolor{Gray}\textbf{29.45}
        &\cellcolor{Gray}\textbf{21.64}
        &\cellcolor{Gray}\textbf{22.13} 
        &\cellcolor{Gray}\textbf{14.79}  
        &\cellcolor{Gray}\textbf{15.22} 
        &\cellcolor{Gray}\textbf{14.06} 
        &\cellcolor{Gray}\textbf{14.37} 
        &\cellcolor{Gray}\textbf{10.95} 
        &\cellcolor{Gray}\textbf{11.68} 
        &\cellcolor{Gray}\textbf{13.76} 
        &\cellcolor{Gray}\textbf{14.54}\\
        
    \midrule

        \multirow{4}{*}{W3A16} & RTN & 1.3e4 & 6.1e3 & 1.6e4 & 1.2e4 & 6.5e3  & 5.8e3 & 4.6e3 & 3.3e3 & 1.5e3 & 1.4e3 & 6.1e3 & 3.6e3\\ 
        & GPTQ \cite{gptq} & 21.17 & 19.45 & 16.83 & 13.75 & 15.09  & 15.67 & 11.73 & 12.28 & 10.30 & 11.34 & 14.42 & 13.68\\ 
        & LSI \cite{lsi} & 19.73 & 19.21 & 16.55 & 14.29 & 14.88  & 15.83 & 11.62 & 12.47 & 10.16 & 11.56 & 14.11 & 14.02\\
        %\cmidrule{2-14}
        &\cellcolor{Gray}\textbf{Ours} 
        &\cellcolor{Gray}\textbf{19.25} 
        &\cellcolor{Gray}\textbf{18.87} 
        &\cellcolor{Gray}\textbf{16.29} 
        &\cellcolor{Gray}\textbf{13.61} 
        &\cellcolor{Gray}\textbf{14.79}  
        &\cellcolor{Gray}\textbf{15.43} 
        &\cellcolor{Gray}\textbf{11.55} 
        &\cellcolor{Gray}\textbf{12.06} 
        &\cellcolor{Gray}\textbf{10.09} 
        &\cellcolor{Gray}\textbf{11.18} 
        &\cellcolor{Gray}\textbf{12.76} 
        &\cellcolor{Gray}\textbf{13.11}\\ 
    \midrule

        \multirow{4}{*}{W4A16} & RTN & 48.17 & 24.68 & 16.92 & 17.61 & 12.10  & 13.38 & 11.32 & 12.35 & 10.97 & 11.90 & 110 & 249\\ 
        & GPTQ \cite{gptq} & 15.56 & 15.57 & 12.82 & 13.75 & 11.41  & 12.15 & 10.31 & 11.36 & 9.63 & 10.80 & 9.55 & 10.50\\ 
        & LSI \cite{lsi} & 15.41 & 15.69 & 12.76 & 14.06 & 11.27  & 12.33 & 10.19 & 11.56 & 9.52 & 11.03 & 9.47 & 10.74\\
        %\cmidrule{2-14}
        &\cellcolor{Gray}\textbf{Ours} 
        &\cellcolor{Gray}\textbf{15.36} 
        &\cellcolor{Gray}\textbf{15.27} 
        &\cellcolor{Gray}\textbf{12.71} 
        &\cellcolor{Gray}\textbf{13.56} 
        &\cellcolor{Gray}\textbf{11.22}  
        &\cellcolor{Gray}\textbf{11.98} 
        &\cellcolor{Gray}\textbf{10.15} 
        &\cellcolor{Gray}\textbf{11.23} 
        &\cellcolor{Gray}\textbf{9.45} 
        &\cellcolor{Gray}\textbf{10.59}
        &\cellcolor{Gray}\textbf{9.43}
        &\cellcolor{Gray}\textbf{10.39}\\ 
    \bottomrule
    \end{tabular}
    \label{tab:opt-w-only}
\end{table*}
%---------------------------------------------------------------------

%------------------------------------------------------------
\begin{table}[tb]
\centering
\footnotesize
%\normalsize
\caption{ Ablation studies on OPT models.}
\label{tab:ablation_study}
%\resizebox{\linewidth}{!}
{
\begin{tabular}{l|l|cc|cc}
\toprule
\multicolumn{2}{l|}{\textbf{Method \& Model}}& \multicolumn{2}{c|}{OPT-1.3B} & \multicolumn{2}{c}{OPT-13B} \\ 
\midrule
- & -  & WIKI & C4  & WIKI & C4 \\ 
\midrule

\multirow{3}{*}{\shortstack{W2A16\\g64}} 
& n=100 & 20.30  &  22.82  & 12.12 & 13.86 \\
& n=50 & 20.97  & 23.11 & 12.24 & 14.06 \\  
& n=200 & 20.15 & 22.69 & 12.06 & 14.11 \\
\midrule

\multirow{3}{*}{\shortstack{W3A16\\g128}} 
& n=100 & 16.21  &  16.29  & 10.24 & 11.42 \\
& n=50 & 16.23  & 16.25 & 10.28 & 11.50 \\  
& n=200 & 16.17 & 16.36 & 10.17 & 11.75 \\
\midrule

\multirow{5}{*}{\shortstack{W4A16\\g128}} 
& - CL & 14.72  &  14.96  & 10.10 & 11.14 \\
& - SDE & 15.26  &  15.63  & 10.25 & 11.44 \\
& LSI & 14.76 & 15.20 & 10.13 & 11.33 \\
& LSI+CL & 15.06 & 15.41 & 10.35 & 11.51 \\
\cmidrule{2-6}
& Full & 14.69 & 14.93 & 10.08 & 11.13 \\

\bottomrule
\end{tabular}
}
\end{table}
%------------------------------------------------------------

\subsection{Other Analysis}
\label{sec:analysis}
\paragraph{Ablation Studies}
Our ablation experiments primarily focus on parameter tuning for SDE and the nuanced effects of both SDE and CL on quantization, as detailed in Table~\ref{tab:ablation_study}. The results indicate that increasing the number of diagonals \( n \) in SDE does not necessarily lead to improved performance. This is because quantization-induced errors cannot be entirely compensated; increasing the number of trainable parameters can cause the model to overemphasize error elimination, which in turn biases the model toward the training data and ultimately leads to degraded performance. On the other hand, SDE itself has strong inherent capabilities for compensating quantization errors, and CL also contributes some enhancements. In addition, applying CL in isolation to LSI \cite{lsi} can even produce adverse effects. This is due to LSI’s extremely small number of training parameters having a disproportionately large impact on the weights, causing LSI to suffer from loss confusion.

%------------------------------------------------------
\begin{figure}[!t]
\includegraphics[width=0.48\textwidth]{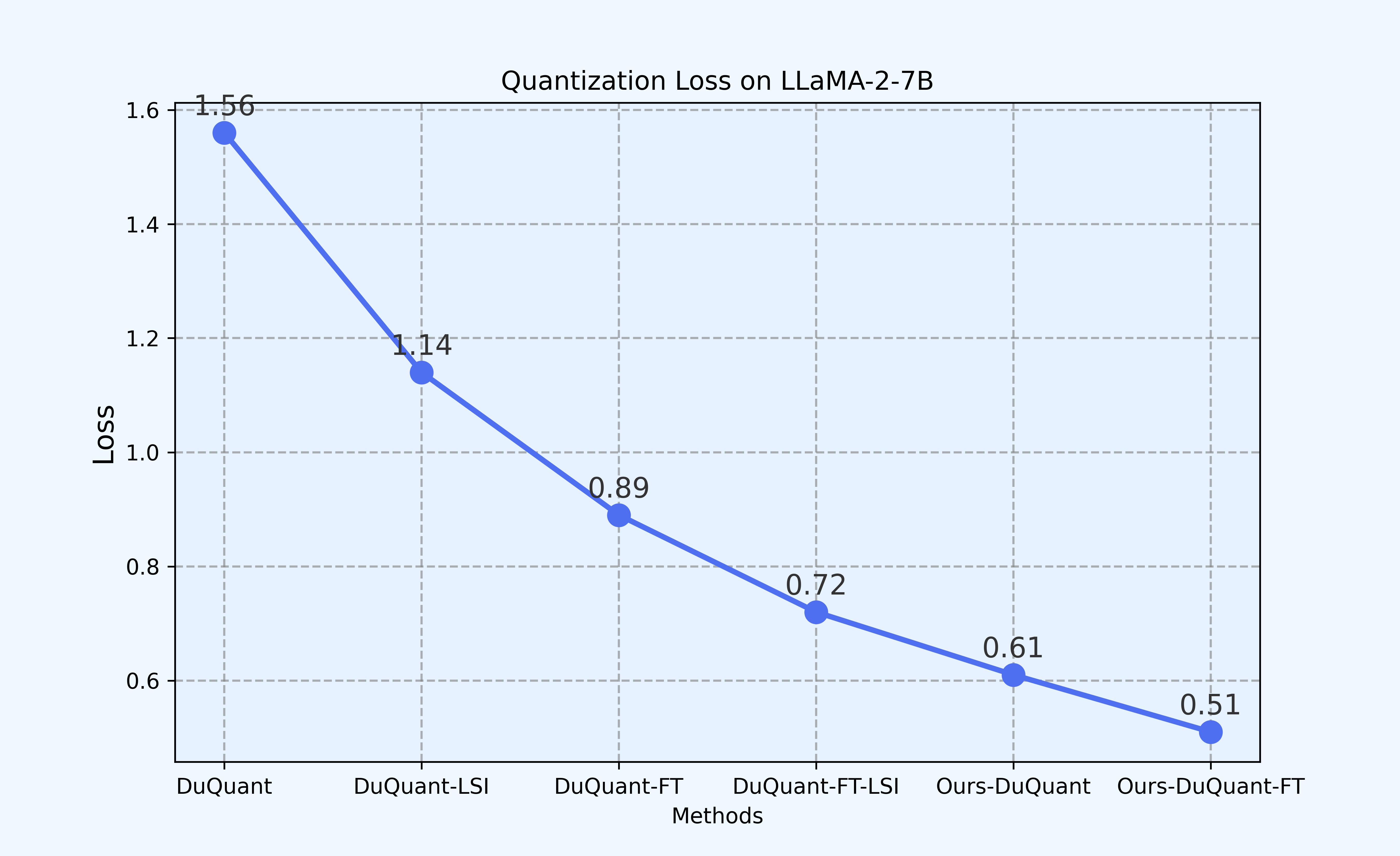}
\centering
%\vspace{-1em}
    \caption{Final quantization loss of different methods on LLaMA-2-7B for DuQuant series.}
    \label{fig:quant_loss_du}
    %\vspace{-1.5em}
\end{figure}
%--------

%------------------------------------------------------
\begin{figure}[!t]
\includegraphics[width=0.4\textwidth]{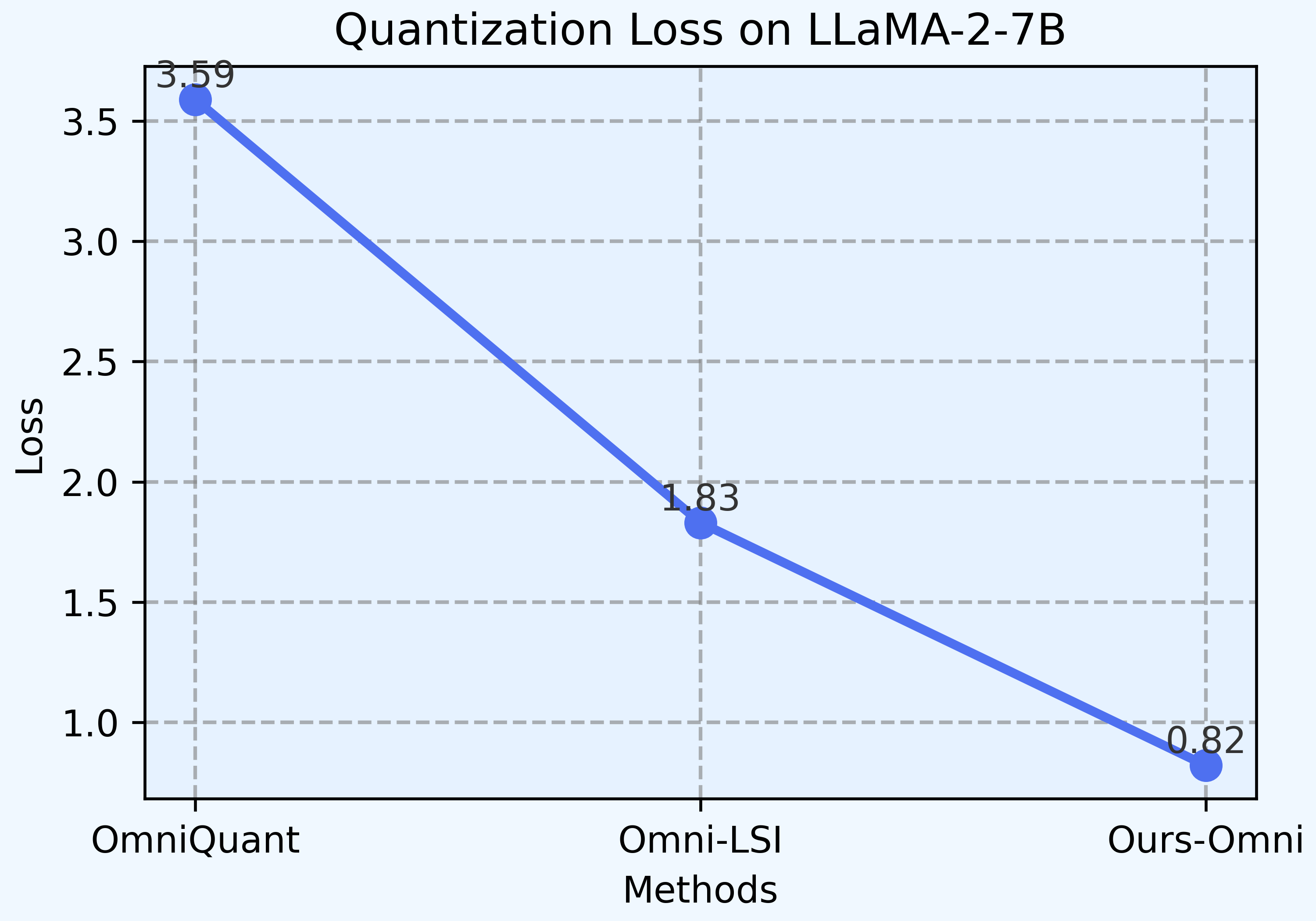}
\centering
%\vspace{-1em}
    \caption{Final quantization loss of different methods on OPT-1.3B for OmniQuant series.}
    \label{fig:quant_loss_omni}
    %\vspace{-1.5em}
\end{figure}
%--------

\paragraph{Overfitting Problem}
Compared to LSI, our approach significantly increases the total number of trainable parameters, amounting to about 2\% of the overall weights. Typically, a larger number of trainable parameters tends to introduce a stronger bias toward the training data—something that is evident in all the baselines. In contrast, our approach shows no corresponding increase in bias. 

Through detailed analysis, we have found that our method is not completely free of bias. The key difference between it and existing weight-quantization methods lies in the way quantization-induced errors dominate. Existing methods are unable to effectively adjust the weights, resulting in poor training data fit and therefore minimal bias. LSI performs better in fitting training data, but its fit is insufficient relative to the quantization errors. As a result, it sacrifices the weight stability and introduces significant bias despite marginally surpassing GPTQ on the training dataset.

In contrast, our approach achieves a much higher degree of fit with the training data. Specifically, as illustrated in Fig~\ref{fig:teaser}(a), our loss is nearly one-third that of GPTQ and half that of LSI for most models, and it becomes even lower under extremely low-bit conditions. This exceptional fit significantly reduces the impact of overall quantization errors, rendering the corresponding bias negligible.

%------------------------------------------------------
\begin{figure}[!t]
\includegraphics[width=0.5\textwidth]{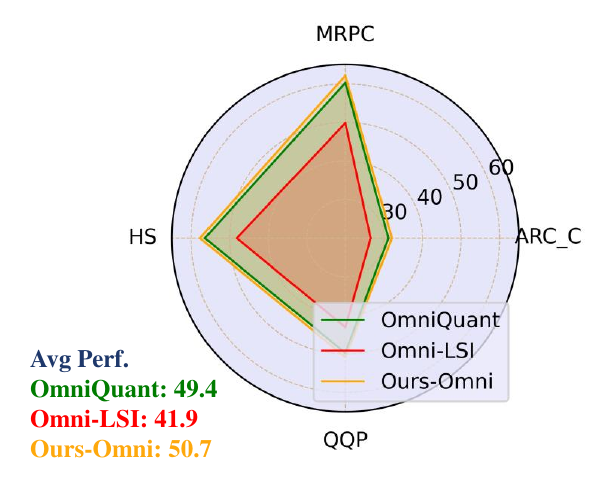}
\centering
%\vspace{-1em}
    \caption{Performance degradation of LSI on LLaMA-7B in W4A4 setting.}
    \label{fig:lsi_bias_perf}
    %\vspace{-1.5em}
\end{figure}
%--------

\paragraph{Training Loss Comparisons}
We also provide a detailed comparison of the final loss values after block-wise calibration across various methods and models, as illustrated in Figs.~\ref{fig:quant_loss_du} and~\ref{fig:quant_loss_omni}. Following calibration using our technique, the final loss is significantly lower than that of other methods~\cite{gptq}\cite{lsi}. For simple DuQuant, which does not involve LWC fine-tuning, our approach reduces the loss to as low as $40\%$ of the original. Even when compared to the loss after incorporating LWC, our method still achieves a reduction of over $40\%$. In the case of OmniQuant, which addresses only normal outliers, the improvement becomes even more pronounced: our final loss is reduced to just $20\%$ of the original baseline. Furthermore, when compared to the loss reduction achieved by LSI on OmniQuant, our method delivers nearly twice the improvement, with final loss values reaching only about half of those produced by LSI.

\paragraph{Bias on LSI}

LSI lacks sufficient degrees of freedom to reliably adjust a large number of parameters, leading to the excessive compromise of some weights for massive outliers, as mentioned previously in Fig.~\ref{fig:weight_disturb} and Sec.~\ref{sec:weight_adjustment}. This approach of significantly ``dragging" weights out of their original distribution levels can lead to substantial degradation in the performance of certain downstream tasks, in some cases even performing worse than not using LSI at all, as shown in Fig.~\ref{fig:lsi_bias_perf}.

\begin{table}[h]
\caption{Comparisons of GPU memory consumption during training.}
\vspace{-10pt}
\normalsize
\label{tab:mem_cost}
\begin{center}

\begin{tabular}{c|c|c}
\toprule
\multicolumn{2}{l|}{\textbf{Model \& Method}} & Memory (GiB) \\ 
\midrule
\multirow{4}{*}{LLaMA-7B} 
& DuQuant+LWC & 10.6  \\
& DuQuant+LSI & 14.8  \\
& DuQuant+SDE & 15.4  \\
& DuQuant+Ours & 17.9  \\

\bottomrule
\end{tabular}
\end{center}
\vspace{-0.7em}
\end{table}

\subsection{Further Discussion}

\paragraph{Inference Speed} 
Our approach does not directly enhance model inference speed. Following training, we can restore the original weights via~\eqref{eq:sde_eq}, reintegrate them into the model, and remove any additional parameters introduced during training. Consequently, under the same quantization configuration, our method maintains inference speeds identical to the original baselines. 

%\input{latex/tables/speedup}

%------------------------------------------------------
\begin{figure}[!t]
\includegraphics[width=0.45\textwidth]{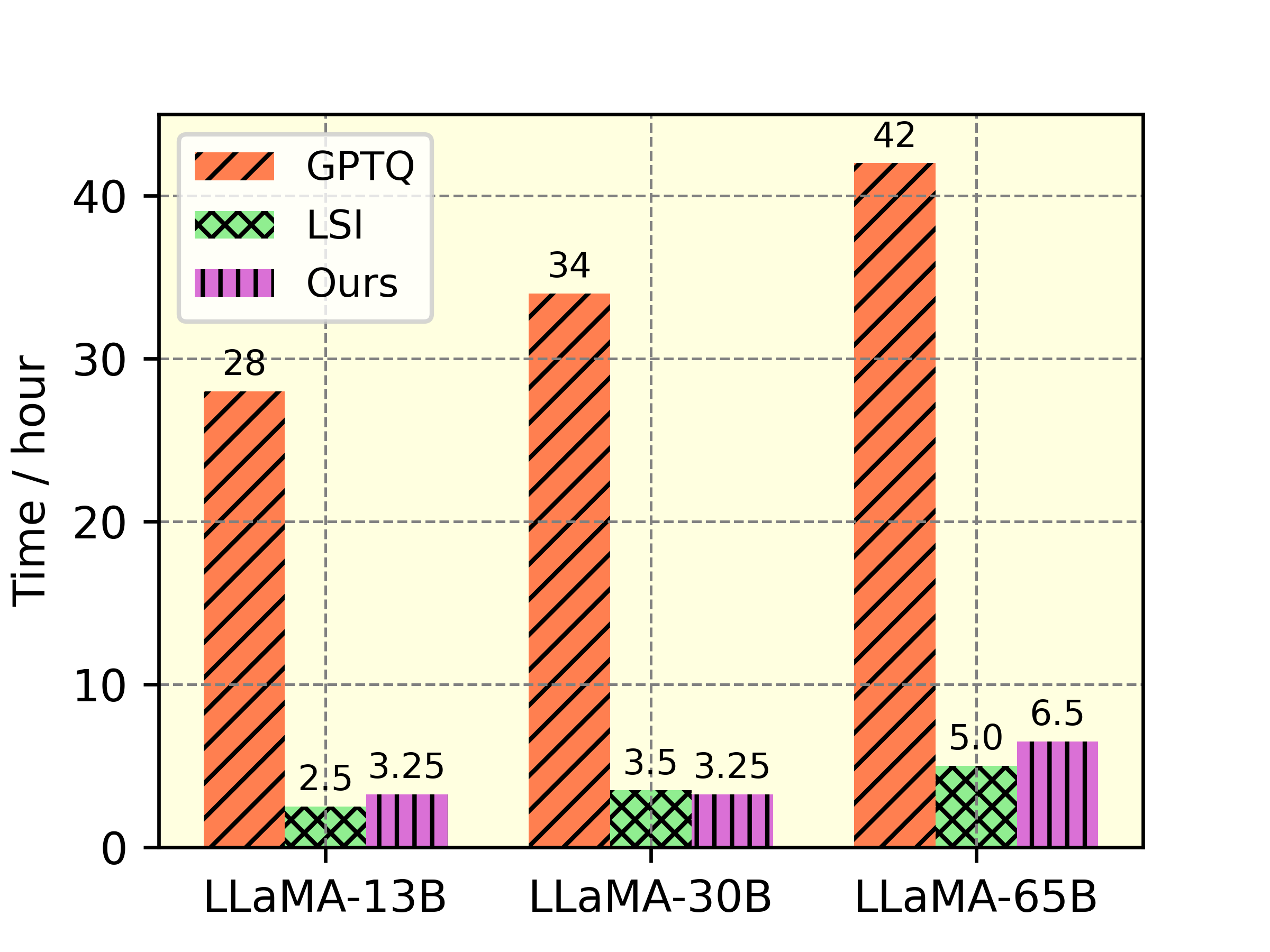}
\centering
%\vspace{-1em}
    \caption{Training time comparisons.}
    \label{fig:train_time}
    %\vspace{-1.5em}
\end{figure}
%------------------------------------------------------

\paragraph{Training Speed and Memory Consumption}
Our training process is fast enough compared with other methods, as shown in Fig.~\ref{fig:train_time}. Its overall time cost is only slightly higher than LSI, yet far below GPTQ’s 24+ hour training duration. The primary computational cost during training arises from the decomposition of SVD matrices. Additionally, the size of \(n\) also impacts the runtime to some extent. When the model size is sufficiently small, a larger \( n \) can reduce the number of required training iterations, resulting in faster convergence—for example, LLaMA-13B with \( n = 200 \) trains faster than that with \( n = 100 \). However, as the model size increases, larger LLMs inherently require fewer training iterations; for instance, LLaMA-65B can converge with just a single iteration using 128 samples. In such cases, an excessively large \( n \) not only introduces additional computational overhead but also risks introducing bias toward the training data.

%------------------------------------------------------
\begin{figure}[!t]
\includegraphics[width=0.4\textwidth]{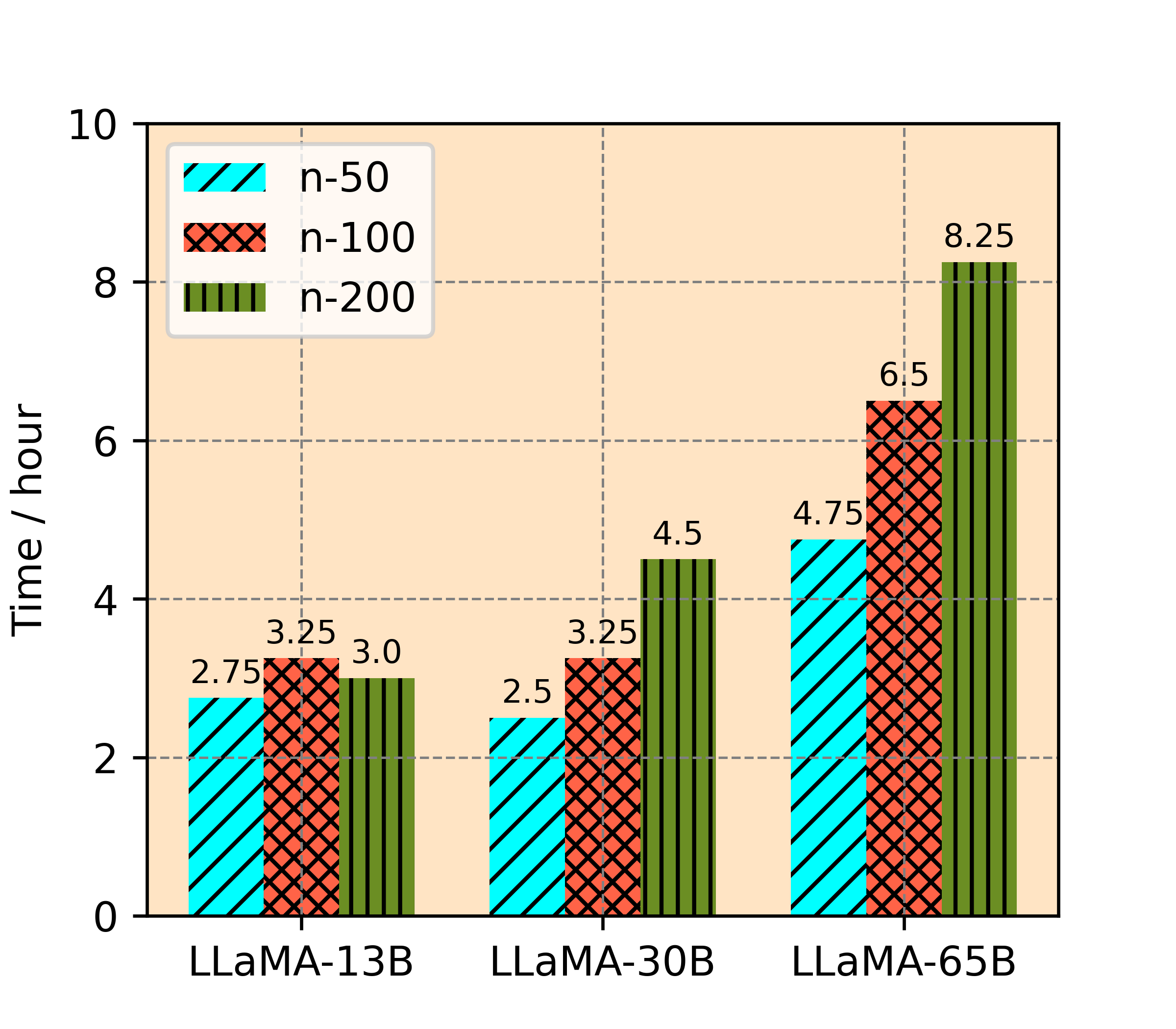}
\centering
%\vspace{-1em}
    \caption{Training time comparisons on different $n$ settings.}
    \label{fig:diff_n_time}
    %\vspace{-1.5em}
\end{figure}
%------------------------------------------------------

We also provide a comparison of training memory usage in Tab.~\ref{tab:mem_cost}, reporting the GPU memory requirements for training LLaMA-7B using DuQuant as the baseline under various methods. The additional training parameters introduced by SDE incur only a marginal increase in memory consumption compared to LSI, whereas CL imposes a relatively higher demand on GPU memory.

\section{Conclusion}
We introduce Singular-value Diagonal Expansion (SDE) to optimize data alignment among various quantization settings, and Cross-layer Learning (CL) to smooth quantization errors across layers. By formulating weight quantization as an inequality-solving problem, we offer valuable insights into attaining globally optimal weight quantization results. Moreover, through theoretical analysis, we demonstrate that SDE offers greater flexibility than LSI in solving inequality-constrained optimization problems, further highlighting the superiority of SDE as a weight quantization method. Our plug-and-play methods enhance the performance of existing state-of-the-art approaches across a range of scenarios. In particular, for low-bit weight-only and weight-activation settings, our techniques not only significantly improve perplexity but also retain nearly lossless downstream task performance. For open-source benchmarks, our method also achieves notable improvements. Furthermore, the training time required by our method is largely lower than that of GPTQ but with much better results, demonstrating its strong potential for industrial application.

\section{Limitations and Future Work}

During training, the additional trainable parameters introduced by SDE do not significantly increase GPU memory usage compared to LSI, as shown in Tab.~\ref{tab:mem_cost}. However, CL requires obtaining the outputs of two consecutive layers, and due to the one-pass nature of gradient backpropagation, we must separately provide two independent inputs—one for the current layer and one for the next layer—substantially increasing the overall parameter count. A practical approach is to store unused tensors on CPU and reduce batch size during training.

During routine loading, if we only store a small number of trainable parameters, reloading the model requires performing an SVD decomposition on the weights, which requires substantial time and presents the same issue as LSI does. Therefore, we recommend saving all quantized model weights after training to avoid the time-consuming process of re-decomposition.

In future work, we plan to explore or develop alternatives to SVD for matrix decomposition—methods that offer faster decomposition and yield submatrices with stronger representational capacity for original matrices. The goal is to reduce training time and enhance model performance in the context of quantization. One promising direction is the use of Principal Component Analysis (PCA) as a potential substitute.

\bibliographystyle{IEEEtran}
\bibliography{main}

\iffalse
\section*{Acknowledgments}
This should be a simple paragraph before the References to thank those individuals and institutions who have supported your work on this article.


\newpage

\section{Biography Section}

\vspace{-33pt}
\begin{IEEEbiography}[{\includegraphics[width=1in,height=1.25in,clip,keepaspectratio]{latex/bio/Yifei_Gao.jpg}}]{Yifei Gao}
Yifei Gao is a master student at University College London. He previously obtained his B.Eng. degree from the College of Software Engineering at the University of Electronic Science and Technology of China in 2024. His current research interests encompass large language models, 3D reconstruction, and diffusion models.
\end{IEEEbiography}

\vspace{11pt}

\vspace{-33pt}
\begin{IEEEbiography}[{\includegraphics[width=1in,height=1.25in,clip,keepaspectratio]{latex/bio/Jie Ou.png}}]{Jie Ou}
Jie Ou is a Ph.D. student at the University of Electronic Science and Technology of China. He previously earned both his B.Eng. and M.Eng. degrees from the University of Electronic Science and Technology of China. His current research interests include 3D reconstruction and accelerating the inference of large language models.
\end{IEEEbiography}
\fi

\vfill

\end{document}